\documentclass[conference]{IEEEtran}
\IEEEoverridecommandlockouts
\usepackage{cite}
\usepackage{amsmath,amssymb,amsfonts}
\usepackage{algorithmic}
\usepackage{graphicx}
\usepackage{textcomp}
\usepackage{xcolor}
\usepackage{lscape}
\usepackage{booktabs}
\usepackage{graphicx}
\usepackage[numbers]{natbib}

\makeatletter
\newcommand*{\rom}[1]{\expandafter\@slowromancap\romannumeral #1@}
\makeatother

\begin{document}

\title{Non-invasive Seizure Detection Using Wearable Wrist-worn Accelerometry and Deep Learning\\
}

\author{\IEEEauthorblockN{Nilushika Udayangani Hewa Dehigahawattage, Kishor Nandakishor and Marimuthu Palaniswami}
\IEEEauthorblockA{\textit{dept. of Electrical and Electronic Engineering } \\
\textit{University of Melbourne}\\
Victoria, Australia \\
}
}

\maketitle

\begin{abstract}
Seizure monitoring and detection are crucial for reducing the morbidity and mortality associated with seizures. Current epilepsy care, often involving expensive video-electroencephalography (VEEG) monitoring, requires specialized expertise and is limited to in-hospital settings, and intrusive in nature. Seizure diaries, on the other hand, suffer from unreliability due to under-reporting, leading to incorrect therapeutic decisions. Wearable non-invasive seizure detection may offer a more tolerable and feasible solution for long-term ambulatory monitoring. This study explores a wearable remote monitoring system utilizing a single wrist-worn accelerometer device and capable of detecting multiple types of seizures, including shorter duration events.
We enrolled 79 patients under video-electroencephalography monitoring to wear accelerometer devices and collect data. Concurrent VEEG recordings were reviewed by board-certified epileptologists to produce annotations, including seizure onset, offset, and seizure type. Using this data, we constructed a deep neural network based on the time-series ResNet architecture, which could discriminate among seizure and non-seizure events.
Our proposed approach achieved a seizure detection sensitivity of 95.65\% and an overall false alarm rate of 0.15/24 hours during the evaluation, which spanned 5576 hours of total recording. Additionally, it resulted in an area under the receiver operating characteristic curve (AUC-ROC) of 0.98 and an area under the precision-recall curve (AUC-PRC) of 0.67 when averaged over 20 patients who experienced 46 convulsive seizures.
These promising results suggest that the proposed seizure detection system can be effectively used for long-term ambulatory seizure monitoring. Future steps include validating our findings in larger datasets and assessing the utility of detection for additional seizure types.

\end{abstract}


\section{Introduction}
\subsection{Background}
\IEEEPARstart{E}{pilepsy} is one of the most prevalent chronic diseases of the nervous system, characterized by recurrent seizures. Globally, approximately 50 million individuals are affected by epilepsy, with 30\% of them diagnosed with intractable epilepsy, a condition showing resistance to anti-epileptic drugs or even multiple drug therapies\cite{50millionGlobally, wenLiReview2022}.
Seizures are caused by abnormal neuronal activity of a population of neuronal cells in the brain. Based upon 1) the seizure onset in the brain (focal onset or generalized onset), 2) the degree of awareness during the seizure (awareness intact or awareness impaired) and 3) the level of body movement (motor or non-motor), seizures are categorized into different types, associated with distinct prognosis and treatment options\cite{seizureTypeClassification}. Focal impaired awareness seizures (FIAS) are the most common known seizure type among adults with epilepsy and generalized tonic-clonic seizures (GTCS) are the most common generalized seizure involving  tonic and clonic convulsions \cite{mostCommonType,commonSeizuresSurvay, keranen_distribution_1988}.

The unpredictable nature of seizures is distressing for the patients and leads to social isolation and decreased quality of life\cite{beniczky_automated_2021}. When unattended or failing to provide urgent medical care, seizures lead to accidental injuries and can evolve into status epileptics, a condition associated with severe complications including neuronal death, neuronal injury and even death \cite{trinkaStatusEpileticus2015}. Nevertheless, the most significant risk associated with seizures is sudden unexpected death in epilepsy (SUDEP), which has a mortality rate of 0.3\% and poses a risk of sudden death more than 20 times higher than that in the general population\cite{shorvon_sudden_2011}. 

The key principle for mitigating seizure-associated risks is early detection and intervention. If the caregivers or physicians could promptly recognize when a patient had a seizure, the injuries/fatalities related to seizures can be potentially reduced. Moreover, it allows physicians to enhance their management of clinical support, diagnosis, and the identification of potential precipitating factors and therapeutic options. Consequently, there is a pressing need for continuous seizure detection and monitoring mechanisms for epileptic patients in both inpatient and outpatient settings.

\subsection{Limitations in current seizure monitoring}
The current options for the detection and monitoring of seizures exhibit limitations on multiple fronts. For instance, video-electroencephalography monitoring (VEM); the current gold standard method for detecting epileptic seizures, and intracranial EEG (iEEG) monitoring, present challenges due to their resource-intensive nature and limited availability, typically confined to tertiary hospitals. This circumstance poses a significant issue, particularly given that approximately 80\% of individuals with epilepsy are situated in low-middle-income countries(LMIC). And partially elucidates why 75\% of patients with epilepsy in LMICs do not receive any form of treatment at all \cite{espinosa-jovel_epidemiological_2018}. Besides VEM, iEEG and ambulatory EEG are obtrusive and not appealing for patients in long-term seizure monitoring. For those reasons, in most cases, long-term continuous seizure recording solely relies on seizure diaries, maintained by patients or caregivers. However, self-reported seizures are unreliable due to under-reporting with less than half of all seizures being accurately documented\cite{LancetDiagnosticChallengesEpilepsy}. The fact that patients are partially or fully amnestic to seizures renders some seizure types including GTCS and FIAS particularly susceptible to under-reporting\cite{lanzone_transient_2018,seizureTypeClassification}. 
This issue is of utmost importance because inaccurate subjective information regarding seizure frequency hinders effective diagnostic and therapeutic decisions and can result in incorrect prescriptions \cite{fisher_bad_2017}.

\subsection{Non-invasive seizure monitoring using wearable devices}
A non-invasive wearable biosensor coupled with a machine learning algorithm has the greatest immediate potential to overcome those barriers \cite{fanghyChenSeizuresDetectionReview2022,brinkmannSeizureDiariesReview2021}. Various recent research has worked toward this direction, employing wearables capable of acquiring non-cerebral physiological signals such as electrocardiography (ECG), electrodermal activity (EDA), surface electromyography (sEMG), accelerometer (ACM), photoplethysmography (PPG), skin temperature, oxygen saturation etc. 

 \subsubsection{seizure detection modalities}
 Several studies employed ECG-derived heart rate variability (HRV) to detect both motor and non-motor seizures \cite{jeppesen_seizure_2019, jeppesen_seizure_2020, giannakakis_focal_2019,jahanbekam_performance_2021,hegarty-craver_cardiac-based_2021}. But most of the works adding patient specificity to their algorithms to improve the seizure detection sensitivity which demands prior recordings of test subjects and  adapting the algorithms on unseen patient data \cite{jeppesen_seizure_2019, jeppesen_detection_2015, van_elmpt_model_2006}. This is primarily due to the substantial variability in cardiac profiles between individuals and within individuals for different seizure semiology, making it difficult to develop generalizable algorithms using ECG derived cardiac data \cite{hegarty-craver_cardiac-based_2021, van_andel_multimodal_2017}. 
Utilizing three distinct and independent datasets, \cite{jahanbekam_performance_2021} further illustrates this fact, highlighting the decline in performance of ECG-based detection methods when training samples and test samples were constructed from different datasets. Another drawback in cardiac activity-based seizure detection is the existence of patients who do not exhibit significant ictal heart rate changes during a seizure, referred to as non-responders in the study \cite{jeppesen_seizure_2020}. Their algorithm yielded a sensitivity of 87\% among responders but exhibited a very low sensitivity of 28\% among non-responders, raising concerns about the universal applicability of HRV-based detection methods.

Given the discomfort associated with wearable ECG electrodes, some studies have explored the use of PPG signals for deriving ictal heart rate \cite{vandecasteele_automated_2017, el_atrache_photoplethysmography_2020, glasstetter_identification_2021}. However, PPG signals alone have shown limited benefit for seizure detection, exhibiting low sensitivities and higher FAR values \cite{vandecasteele_automated_2017, glasstetter_identification_2021}. The fact that PPG is easily interfered by motion artifacts is known to be the main reason behind the poor performance \cite{wenLiReview2022}. To enhance sensitivity, the use of multimodal approaches, such as coupling PPG with ACM, is demonstrated \cite{arends_multimodal_2018}.

Some researchers have also demonstrated the utility of sEMG for detecting seizures \cite{szabo, conradsen, beniczky2018, baumgartner_using_2021}. The majority of sEMG-based studies have reported high specificity values in their performance. Moreover, compared to algorithms triggered by clonic jerks, sEMG-based algorithms have demonstrated shorter GTCS detection latency, as they target the tonic phase \cite{beniczky_non-electroencephalography-based_2019}.
However, the surfaces of the sEMG sensors require continuous and strong attachment to the skin and have the potential for detachment, which can lead to patient discomfort. Additionally, the location of sEMG acquisition significantly impacts the algorithm's performance\cite{wenLiReview2022, ulate-campos_automated_2016, conradsen}. These factors limit the use of sEMG for long-term seizure detection.

Electrodermal activity (EDA), which reflects the activity of the sympathetic branch of the autonomic nervous system, has also been identified as a potential seizure biomarker in recent studies \cite{poh_continuous_2010, poh_autonomic_2012}. Importantly, a significant increase in EDA has been associated with SUDEP, highlighting the potential of EDA in mitigating SUDEP risk \cite{poh_autonomic_2012}.
However, there are no studies that advocate for EDA as a single modality, instead, EDA is commonly utilized in multimodal studies \cite{fanghyChenSeizuresDetectionReview2022,wenLiReview2022}. In a study \cite{tangJianbin2021}, comparing different modalities, EDA did not achieve significant better than chance seizure detection, as indicated by overall AUC-ROC, whereas other modalities did. Due to the fact that skin conductance response is relatively slow, 
EDA shows a delayed response time \cite{tangJianbin2021}.  
This latency in the onset of EDA might be the main underlying reason why EDA fails as a single modality but performs well when combined with other modalities. Other disadvantages associated with EDA include its susceptibility to motion and pressure artifacts, as well as its uncomfortable and obtrusive nature \cite{ulate-campos_automated_2016}.

Accelerometry (ACM) presents a viable alternative modality to address these challenges, offering the advantage of being worn as a comfortable wristband-type device. Since ACM measures the three-dimensional acceleration of movements and rotations, it can be employed for detecting seizures based on their clinical motor symptom, one of the main criteria for diagnosing a seizure. Furthermore, ACM's ubiquity in consumer electronics, including smartwatches, smartphones, and fitness monitors, has contributed to its widespread acceptance and familiarity among individuals. Notably, recent findings indicate that smartphones and watch-based devices were acceptable to over 70\% of people with epilepsy, in contrast to devices placed on the head, leg, upper arm, and chest, which garnered an acceptance rate of $<$50\%
\cite{bruno_wearable_2018}. Moreover, its cost-effectiveness and low power consumption are among other advantages of ACM. This underscores the favourable reception and suitability of ACM for long-term seizure monitoring. Recent studies have demonstrated its utility in detecting motor seizures  \cite{agrahriAnushman2022, baseline, kramer, lockman}.  One study revealed that, when compared to other modalities ACC achieved the best performance for most seizure types and improved the performance of other modalities when coupled with ACM \cite{tangJianbin2021}. In a result analysis among 30 studies \cite{fanghyChenSeizuresDetectionReview2022},  many of the reported highest sensitivities and lowest FARs were associated with the incorporation of ACM as a sensor modality.  

Seizure-related changes in physiological signals vary across different seizure semiologies, making some sensor modalities more applicable to specific seizure types. For instance, the effectiveness of ACM and sEMG is better demonstrated in motor seizures, whereas their utility is limited in seizures lacking apparent motor symptoms. Conversely, ECG and PPG can be employed for non-motor seizures, albeit with reduced sensitivity to motor seizures \cite{wenLiReview2022}. 
As a result, certain studies have adopted multi-modality systems to address a wide range of seizure types. Commonly used combinations include ACM with EDA \cite{onorati, poh2012,heldberg_using_2015}, ACM with sEMG \cite{gheryani_epileptic_2020,milosevic2016} and ECG with ACM \cite{munch_nielsen_towards_2022,van_andel_multimodal_2017}). Furthermore, certain studies have employed multiple sensors from the same modality to ensure better sensor placement on the body for optimal seizure registration and detection, enhance signal quality, and reduce missing data and issues aroused due to sensor failures \cite{milosevic2016, johansson_tonic-clonic_2019}. 
While studies involving multiple modalities and multiple sensors have shown promise, their practical application in real-world scenarios is constrained by several factors. These limitations include burdens related to instrument portability, extensive data handling and processing requirements, elevated power consumption leading to limited expected battery life, synchronization issues among multiple signals, and obstacles in terms of commercialization and clinical implementation. Consequently, still there is significant potential for research in epileptic seizure detection using single-modal signals and a minimal number of sensors, possibly even a single sensor. Such an approach holds substantial research prospects, especially if it can successfully detect multiple seizure types, which would be extra advantageous.

\subsubsection{seizure detection algorithms}
seizure detection algorithms are integral to seizure detection with wearable biosensors. The algorithms used in recent studies can be mainly identified in three categories, traditional machine learning(SVM, kNN, random forest, decision tree etc), deep learning(1D CNN, LSTM) and other algorithms including thresholding and event searching \cite{fanghyChenSeizuresDetectionReview2022}. 
The most commonly used algorithms were threshold-based, where the optimal value is being picked as threshold to pass the event as an epileptic seizure. The rise of heart rate \cite{hegarty-craver_cardiac-based_2021}, number of zero crossings\cite{beniczky2018, conradsen} and z-score \cite{gheryani_epileptic_2020,hegarty-craver_cardiac-based_2021} were among the parameters that threshold were derived from. But those methods inherently carries drawbacks associated with threshold-based classification such as not robust to signal noise, poor generalization on new or unseen data,  lack of context and less adaptability under evolving data characteristics of a patient and variability across patients.

In contrast machine learning algorithms allow the systems to adaptively learn patterns in data, which may not be apparent to human observer.  
Among traditional machine learning methods, SVM \cite{poh2012, milosevic2016, vandecasteele_automated_2017,onorati_multicenter_2017} was the most commonly employed and random forest \cite{dongChunjiao2021, forooghifar_self-aware_2022, vandecasteele_power_2021} , k-Nearest Neighbours \cite{johansson_tonic-clonic_2019, heldberg_using_2015}, gradient tree boosting \cite{bottcher_intra-_2022, bottcher_detecting_2021} were among rest of the commonly utilised methods. 
However, these methods involve laborious feature engineering steps, where raw data is preprocessed, and time domain features, frequency domain features, or other characteristics of interest are extracted. These features are selected for their discriminative power in distinguishing between seizure and non-seizure events and among different types of seizures. Hand-crafted features may struggle to capture complex and nonlinear relationships within the data, potentially missing intricate patterns and relationships, which can limit the model's representational power. Besides these features are subjective to the particular study, the dataset being studied and targeted seizure types. 

In contrast deep learning methods provides "end-to-end" learning , where the salient features extracted through initial layers enables algorithms to be seamlessly tailored to a single seizure type or multiple seizure types. Several recent studies have employed deep learning networks as seizure detection algorithm, including 1D CNN \cite{tangJianbin2021}, LSTM \cite{meisel_machine_2020, nasseri_non-invasive_2021}, variational autoencoder (VAE) \cite{you_semi-supervised_2022}, generative adversarial network (GAN) \cite{you_unsupervised_2020}, and residual 1D CNN (Res1DCNN) \cite{baghersalimi_personalized_2022}. The VAE and GAN developed using behind-the-ear EEG reported higher sensitivities ($>$90\%) and lower FARs ($<$1d$^-1$), which were comparable to traditional machine learning methods. However with reported sensitivities below 80\% or FARs around 10d$^-1$, results looked less promising, in LSTM and CNN networks developed using EDA, ACM, BVP, PPG and TEMP \cite{meisel_machine_2020, nasseri_non-invasive_2021, tangJianbin2021} indicating certainly existing opportunities for further enhancement in performance of seizure detection tasks using deep learning. 
Deep learning is better equipped to learn complex relationships and variabilities among different modality signals and across various channels within single-modality multivariate signals. However, most existing deep learning networks primarily emphasize multimodal signals. Further exploration of their applicability to single-modal signals in the multivariate analysis perspective is warranted.

Seizure detection algorithms can be identified in two categories namely personalized and generalized. The distinction between them lies in whether the algorithm is designed for a specific patient or can be applied across any patient (including the unseen during training). Personalized algorithms which tailored to a specific patient, often reported high sensitivities and low FARs compared to generalized counterparts \cite{ulate-campos_automated_2016}. However, they are costly in terms of data acquisition, as they require collecting patient-specific training data for each individual patient and tuning the algorithm separately. This poses challenges for commercialization, reduces practicability, and limits clinical utility. On the other hand, generalized algorithms could reduce these costs by developing models that can be applied across patients. But the inherent variability among patients in terms of clinical symptoms and pathophysiological conditions related to seizures makes the implementation of generalized algorithms challenging. The key to addressing this challenge is to only focus on features that are not influenced by individual patient characteristics. Deep learning algorithms are well-suited to this task, as they can automatically learn relevant features that are generally applicable and evolving nothing about individuals \cite{wenLiReview2022}.

Seizures represent exceedingly rare events, constituting an exceptionally small fraction of long-term monitoring data in comparison to non-seizure intervals corresponding to normal behaviour. Therefore, the issue of class imbalance should be a pivotal consideration in the design of seizure detection algorithms. This imbalance becomes more pronounced when the minimum targeted seizure duration is reduced to smaller values.
Furthermore, it is worth noting that many studies exclude patient data in which no seizures occurred during the recording period. Consequently, the degree of imbalance encountered in real-world patient data far exceeds that encountered during the training phase. As a result, algorithms that give minimal attention to imbalance may not perform well in real-world seizure detection. Despite this, only a few studies have taken class imbalance into account during algorithm development. Among these, some have employed novelty detection or anomaly detection-based methods \cite{yamakawaWear2020Anomaly, you_semi-supervised_2022, you_unsupervised_2020,cuppens_accelerometry-based_2014}. Additionally, the support vector data description (SVDD) method \cite{baseline} has been used to address class imbalance. However, all of these proposed methods exclusively adopt unsupervised learning approaches, omitting the potential benefits of supervised learning algorithms, which could offer better control in achieving the desired balance between sensitivity and specificity.

\subsubsection{limitations in current ACM-based seizure detection and  monitoring}
Eventhough ACM is the dominant modality in non-invasive seizure detection, current studies limited in multiple senses. Since ACM focus on motor behaviour, rhythmic movements in daily life such as brushing teeth would also mimic the seizure related convulsions or jerking movements. Therefore algorithms that not capture these intricacies in data may lead to lots of false alarms. 
It has been repeatedly reported that motivation to use wearables by patients will depend on the accuracy and reliability of the device to detect or predict seizures, with a low false positive/false negatives rates\cite{simblett_patients_2020}. Therefore reducing FAR is a crucial factor and most of the existing literature is yet to achieve a fairly acceptable FAR. 
Most of the ACM-based seizure detection algorithms only focus on detecting GTCS, which exhibit prominent motor activity but ignore other seizure types like focal seizures.


\begin{table}[t]
\centering
\caption{Seizure distribution among patients.}
\label{table:overview_of_patients}

\scriptsize
\setlength{\tabcolsep}{7pt}
\renewcommand{\arraystretch}{1.05}

\begin{tabular}{@{}cccc@{}}
\toprule
\textbf{Patient ID} & \textbf{\#GTCS} & \textbf{\#PNES} & \textbf{\#FIAS} \\
\midrule
$P_1$    & 2 & 0 & 0 \\
$P_2$    & 2 & 1 & 0 \\
$P_3$    & 1 & 0 & 0 \\
$P_4$    & 1 & 0 & 0 \\
$P_5$    & 2 & 0 & 0 \\
$P_6$    & 1 & 0 & 0 \\
$P_7$    & 1 & 0 & 0 \\
$P_8$    & 1 & 0 & 0 \\
$P_9$    & 1 & 0 & 0 \\
$P_{10}$ & 1 & 0 & 0 \\
$P_{11}$ & 6 & 0 & 0 \\
$P_{12}$ & 2 & 0 & 1 \\
$P_{13}$ & 0 & 2 & 0 \\
$P_{14}$ & 0 & 5 & 0 \\
$P_{15}$ & 0 & 1 & 0 \\
$P_{16}$ & 0 & 1 & 0 \\
$P_{17}$ & 0 & 9 & 0 \\
$P_{18}$ & 0 & 1 & 0 \\
$P_{19}$ & 0 & 0 & 1 \\
$P_{20}$ & 0 & 0 & 3 \\
\midrule
\textbf{Total (20 patients)} & \textbf{21} & \textbf{20} & \textbf{5} \\
\bottomrule
\end{tabular}
\end{table}
\section{Methods}
\subsection{Data Acquisition}
The data used in this study were collected from patients undergoing long-term VEM at Royal Melbourne Hospital. To capture motor activity data, Apple iPod Touch devices equipped with built-in microelectromechanical systems (MEMS) accelerometer sensors were employed. These devices were securely strapped to the dorsal side of the monitored patients' forearms. A custom-designed mobile application was developed to continuously record 3D ACM data, which had a sensitivity range of -2.5g to +2.5g. 
The recorded data, along with corresponding timestamps, were stored in the flash memory of the devices and subsequently sampled at a rate of 50 Hz. To enable synchronization with the VEM recordings, the time settings were manually updated. Each recording session had a duration of 12 hours, and the devices were replaced with fully charged ones to ensure continuous coverage throughout the entire VEM period. It is noteworthy that the patients were not confined to their beds during the recording sessions and were encouraged to engage in their usual daily activities. Ethical approval for the study was obtained from the Human Research Ethics Committee of the Royal Melbourne Hospital (HREC Project 300:259).

\subsection{Recorded Dataset}
The data collection period spanned from 2012 to 2015 and included records from a total of 79 patients, resulting in a cumulative recording duration of 5576 hours. Each seizure event was meticulously reviewed by expert epileptologists to determine its type, onset, offset, seizure semiology as observed on VEM, and localization. Importantly, the epileptologists conducting the reviews were blinded to the accelerometer traces. Out of the 79 patients, 20 (25.31\%) experienced convulsive seizures, 15 (18.98\%) had non-convulsive seizures, while 44 patients did not exhibit any seizures during the monitoring period. This study focuses exclusively on convulsive seizures. The 20 patients who had convulsive seizures accounted for approximately 1397 h of recording duration. Among them a total of 46 convulsive seizures were observed comprising 21 GTCS, 20 PNES, and 5 FIAS. As shown in table \ref{table:overview_of_patients} ,
among the 20 patients , 10 (50\%) had only GTCS events, 6 (30\%) had only PNES events, and 2 (10\%) had only FIAS events. Additionally, 1 patient each experienced both GTCS + PNES and GTCS + FIAS events.

To provide context on the seizure types includes in the dataset:
\begin{itemize}
\item GTCS is characterized by abnormal electrical activity originating simultaneously in both the right and left cerebral hemispheres, often leading to impaired awareness or complete loss of consciousness. It typically consists of a tonic (stiffening) phase followed by a clonic (jerking) phase.
\item FIAS occurs when abnormal electrical activity starts in one side of the cerebral hemisphere and is associated with memory loss and jerking of a single limb. 
\item PNES is characterized by events that do not result from epileptiform discharges in the brain yet superficially resembles epileptic seizures. These events typically involve impairment of consciousness, flaccid or rigid collapse, and/or tremulous limb movements \cite{brown_psychological_2016}.
\end{itemize}

Given that FIAS is the most common seizure type, GTCS is the most prevalent among generalized seizures, and PNES is the most common non-epileptic seizure type, the inclusion of these three seizure types allows our dataset to represent a substantial portion of the patient cohort experiencing seizures.

The distribution of the observed convulsive seizures, categorized by the minimum seizure duration, is presented in Table \ref{table: seizure_duration_distribution}. Notably, the number of observed seizures decreases as the minimum seizure duration increases. To gain a more comprehensive insight into the utility of our algorithm concerning specific motor behaviors associated with different seizure types, we organized our data into four distinct groups. This categorization follows the same methodology outlined in our baseline work \cite{baseline}. The four groups are as follows: 1) PG1: Patients with only GTCS, 2) PG2: Patients with GTCS and FIAS, 3) PG3: Patients with GTCS and PNES, and 4) PG4: Patients with GTCS, PNES, and FIAS.

\begin{table}[]
\caption{Seizure Durations Distribution}
\label{table:seizure_duration_distribution}
\resizebox{\columnwidth}{!}{%
\begin{tabular}{ccccccc}
\hline \hline
seizure duration & $\geq 10s$ & $\geq 20s$ & $\geq 30s$ & $\geq 40s$ & $\geq 50s$ & $\geq 60s$ \\ \hline
\#patients & 20/79 & 20/79 & 18/79 & 18/79 & 17/79 & 14/79 \\
\#seizures & 46/46 & 45/46 & 39/46 & 34/46 & 29/46 & 25/46 \\\hline \hline
\end{tabular}%
}
\end{table}

\subsection{Pre-Processing}
We adopted same pre-processing steps as described in the baseline work \cite{baseline} which ACM data is formed into so called "events" (seizure events and non-seizure events).
As the first step the resultant signals of 3D ACM data was generated using \(R = \sqrt{a_x^2 + a_y^2 + a_z^2}\). Let (\ref{discrete_time}) represent the resultant signal as a discrete time signal. 
\begin{equation}
    x[n] = x(nT) = \{x_1, x_2, x_3, ...\} \label{discrete_time}
\end{equation}
where \(n \in Z\) and T is the sampling period.
The ACM signal was analyzed using 10s windows with 50\% overlap. The choice of window size of 10s is made based on the minimum seizure duration (10s) considering in the study. The first stage of pre-processing involves activity filtering, where it selectively filters out segments with no activity or subtle activity. The activity filtering starts with thresholding the signal at 0.2 g and can be summarized in (\ref{thresholding}). The value of 0.2 g is empirically chosen based on the lower bound of the collected seizure data.
\begin{equation}\label{thresholding}
    x_t[t] =  \begin{cases} 
            x[t], & \text{if \(|x[t]| \geq 0.2 g\)} \\
            0, & \text{otherwise}
            \end{cases}
\end{equation}
where \(x_t[t]\) denotes the thresholded signal. Followed by the thresholding step, signal is filtered using a sixth-order Butterworth bandpass filter with the intention of filtering out spurious spikes. The cutoff frequencies were set to 2 and 25 Hz frequency range based on the observations of dominant frequency of typical seizure events from the previous works \cite{bayly_time-frequency_2013,gubbi_automatic_2016}.  
The last step of activity filtering involves evaluating activity measure per each window using:
\begin{equation}\label{activity}
activity_n = \frac{1}{(10 \times F_s)} \sum_{t=a}^{b} |sgn(x'[t])|
\end{equation}
where, \(activity_n\) represent the activity of \(n\)th window, \(F_s\) is the sampling rate, \(a = (n-1) \times \frac{W_{len}}{2} \times F_s + 1\), \(b = (n + 1) \times \frac{W_{len}}{2} \times F_s\), \(W_{len} = 10\) and \(sgn(x)\) is the signum function defined by:
\begin{equation}\label{signum}
sgn(x) = \begin{cases}
            1, & x \geq 0 \\
            - 1, & x < 0
            \end{cases}
\end{equation}.
The windows with more than 50\% activity is identified as active windows with sustained activity and subsequently merged into one if they are lies within \(< 20s\) gap. This merged signals are termed here as 'events'. They include both non-seizure events(ADL) and seizure events that vary in duration. The next stage of the preprocessing is time-filtering where it filters out all the events that having duration $<10 s$ (minimum seizure duration). As the last step we resize the each event to have same number of samples(500) using upsampling/downsampling in Fourier method.
As a whole, preprcessing steps segment the whole data of each patient into several events that having duration $>10 s$ and 500 time-steps. The total number of events corresponding to 79 patinets were 8362 among which 46 seizure events and 8316 of non-seizure events.

This way, pre-processing steps contributed to reduce the volume of the data and mitigating the class imbalance problem for some extent, and computational complexity as well.

\begin{figure}[htbp]
\centerline{\includegraphics[scale=0.16]{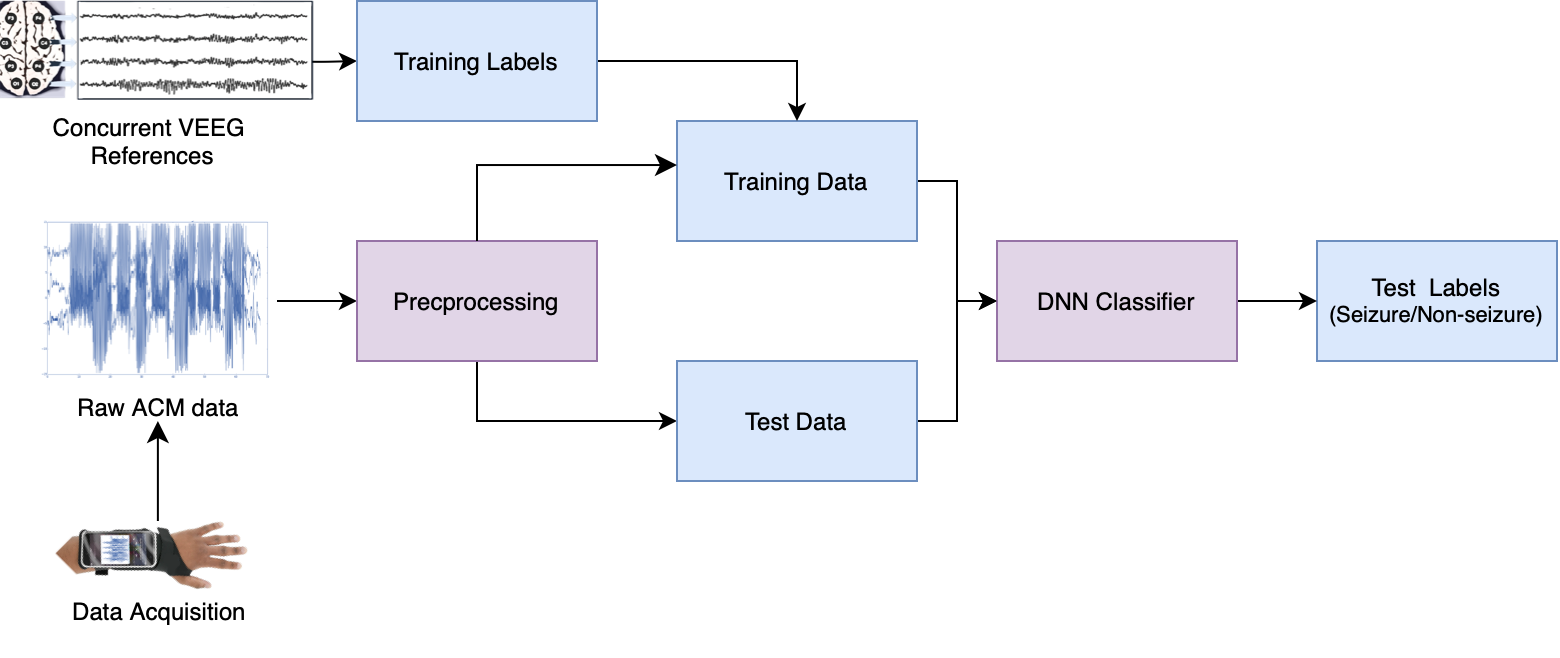}}
\caption{Overview of the Proposed Method.}
\label{figure:method}
\end{figure}

\subsection{Time Series Residual Network}
We constructed a deep neural network following the Residual Network architecture proposed in \cite{wang_time_2017}. In plain networks, when backpropagating over large number of weight layers gradient can decrease exponentially quickly to zero termed as vanishing gradients problem, which hamper convergence from the beginning. Another issue is even though network begin to converge, with the network depth increasing, it generate higher training error that is irreducible even after large number of iterations, termed as network degradation problem. Therefore making a plain networks deeper may either end up with hindering their convergence or degrading the performance. ResNet alleviate theses two problems using shortcut connections and residual blocks allowing to construct very deep neural networks to learn complex data. In backward propagation shortcut connections enable the gradient flow directly through the bottom layers alleviating the gradient vanishing problem and in forward propagation, input signals can be directly transmitted from any low level to high level reducing the degradation problem \cite{he_deep_2016,sun_low-degree_2022}.  It was originally developed for object detection and other vision-related tasks and achieved milestone performance with a relative improvement of 28\% for some benchmark datasets \cite{he_deep_2016}. Its application to time series classification later being explored \cite{wang_time_2017} and achieved state-of-the-art performance when evaluated on UCR time-series archive \cite{chen_ucr_2015}.

\begin{figure}[htbp]
\centerline{\includegraphics[scale=0.38]{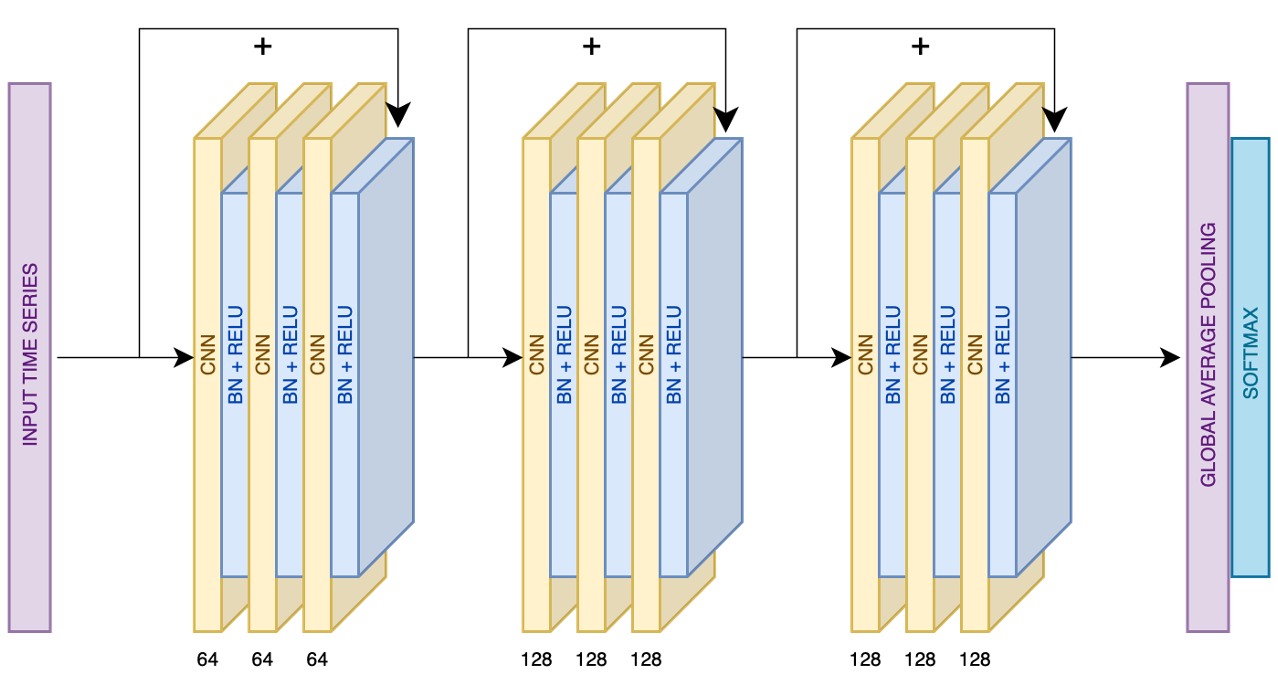}}
\caption{Time Series ResNet Architecture.}
\label{figure:resent_architecture}
\end{figure}

The implemented version of ResNet includes three residual blocks as depicted in Figure \ref{figure:resent_architecture}. Each residual block is first composed of three stacked convolutional blocks whose output is added via a shortcut connection to the same residual block's input and then fed to the next residual block. The basic convolutional block consists of a convolutional layer followed by a batch normalization (BN) layer and a rectified linear unit (ReLU) activation layer and can be represented in:
\begin{align*}\label{conv_block}
y = \textbf{W} \otimes \textbf{x} + \textbf{b}\\
s = BN(y)\\
h = ReLU(s)\tag{1}
\end{align*}
where $\otimes$ is the convolution operator. The ReLU activation function, handles the non-linearity while prevent the gradients from being saturated when the network becomes deep. Batch normalization on the other hand contributes to speedy convergence and improved generalization. To prevent overfitting , pooling layers has omitted in the design. If $CB_i$ denotes the convolutional block, then the residual block can be formulated as:
\begin{align*}
h_1 = CB_1(x)\\
h_2 = CB_2(h_1)\\
h_3 = CB_2(h_2)\\
y = h_3 + x\\
h = ReLU(y)\tag{2}
\end{align*}
In each residual block the filter lengths are set to 7, 5 and 3 respectively for the first, second and third convolution. All convolutions in the first residual block are performed with 64 filters and it is 128 for the rest of the residual blocks. Followed by the residual blocks is a global average pooling (GAP) layer and a final softmax layer whose number of neurons equals the number of classes('seizure' or 'non-seizure'). It is important to note that feeding final feature maps in to a GAP layer instead of a fully connected layer heavily reducing the number of weights. 

\subsection{Training DNN}
Under class imbalance, a model is exposed to a significantly larger number of negative samples than positive samples during training. This makes the model learn to predict more negatives than positives since minimizing the training loss is favored by doing so. To address this issue, class weights can be introduced to modify the loss function, preventing the model from taking the "easy solution" of predicting negatives. In our study, we employed weighted binary cross-entropy loss function, denoted in (3), with prediction $\hat p$, label $p$, and hyperparameter $\beta$ defining the positive class weight. By finetuning this hyperparameter, we can adjust the balance between the reward for true positives and true negatives.
\begin{equation*}WBCE\left( {p,\hat p} \right) = - \left( {\beta p\log (\hat p) + (1 - p)\log (1 - \hat p)} \right)\tag{3}\end{equation*}
When $\beta > 1$, a higher value gets multiplied with the negative log-likelihood of the true class, effectively up-weighting the loss for the underrepresented class and imposing a higher penalty. In our training process, we experimented with various values of $\beta$ (all $>1$), ultimately determining that a value of 3 yielded the optimum setting.

For training the ResNet, we employed the 1cycle policy proposed in \cite{smith_super-convergence_2018}. Training a Deep Neural Network (DNN) poses a challenging global optimization problem, with the learning rate (LR) being a critical hyperparameter to fine-tune during training. Ideally, one should start with an optimal LR and gradually decrease it towards the end. Traditional LR schedulers, such as Time-Based Decay, Step Decay, Exponential Decay, and others, are set before training and remain constant throughout. In contrast, adaptive LR methods like AdaGrad \cite{duchi_adaptive_nodate}, Adam \cite{kingma_adam_2014}, AdaDelta \cite{zeiler_adadelta_2012}, and others offer locally adaptive LRs but come at a higher computational cost.
Cyclical Learning Rates (CLR), introduced in \cite{smith_cyclical_2017}, provide a computationally efficient alternative to adaptive LR methods. CLR allows the LR to cyclically change between minimum and maximum boundaries instead of following a fixed or exponentially decreasing schedule. Each CLR cycle consists of two phases: one in which the LR increases and another in which it decreases. The optimal LR is expected to fall between the minimum and maximum bounds. Building upon the CLR method, \cite{smith_super-convergence_2018} introduced the 1cycle policy, which enables DNN training with an order of magnitude greater speed than standard methods.
The 1cycle policy employs the CLR method but with just one cycle and a notably large maximum LR bound. It starts with a low LR, increases it to a significantly high value, and then decreases it to a value much lower than the initial LR. In this approach large LR serves as a regularization technique, keep network from overfitting, greatly enhancing training efficiency and achieving higher test accuracy compared to standard training protocols. In particular, it has been demonstrated with shallow ResNet architectures that the 1cycle policy leads to a notable increase in accuracy. Importantly, the benefits of this method become even more pronounced when dealing with a limited number of labeled training samples \cite{smith_super-convergence_2018}. These observations, which align with our specific use case of employing a shallow ResNet architecture with a limited available training examples, provide strong motivation for the adoption of the 1cycle policy.

We determined the maximum LR for the 1cycle policy through an LR range test \cite{smith_cyclical_2017}. This test involves a pre-training run in which training begins with a small LR and is gradually increased linearly after each mini-batch until the loss value starts to explode. The LR value at this extremum point, where the loss is at its minimum before starting to explode, is the largest value for the maximum bound for LR. We selected a value between this point and the point with the steepest slope. Additionally, to prevent overfitting, we employed early stopping based on the validation loss with a patience of 3 epochs. All the analyses were performed with a written code using Python and tsai (timeseriesAI) library.

\subsection{Datasets Formation and Experimental Setup}
To leverage the available limited data optimally for test result generation and to present these results in a generic, non-patient-specific manner, we adopted the leave-one-patient-out (LOPO) cross-validation approach. The data from all patients who experienced at least one seizure during the monitoring period were included in the training process of the LOPO models. There were 12 such patients for patient group PG1, while there were 14, 18, and 20 such patients for groups PG2, PG3, and PG4, respectively. Conversely, data from patients who did not experience any seizures were employed as an independent population to assess the generalization of the developed models, in terms of false alarm rate (FAR).
Let $P_{s\_all}$ denote the set of patients who had seizures and let $P_{ns\_all}$ represent the patients without any seizures. Furthermore, let $P_i$ designate the patient whose data were excluded from the training phase of the specific LOPO model, denoted as $LOPO_i$. Then the training set and test set for the $LOPO_i$ model consist of data from distinct patient sets: \{$P_{s\_all} - P_i$\} and \{$P_i$\} respectively. Additionally, the training set is further subdivided such that two-thirds of it comprises the learning set, while one-third is allocated to the validation set. Importantly, no shuffling of data across patients is performed during this splitting.
Each $LOPO_i$ model undergoes ten separate training iterations with different randomizations to select the best-performing model. The criteria, determining the best-performing model will be described later.

To assess the generalization of the proposed method by evaluating the FAR on non-seizure patients, another four separate models were trained (one per each group PG1-PG4). In each of these models, the training set comprised corresponding \{$P_{s\_all}$\}, while the test set comprised corresponding \{$P_{ns\_all}$\}. The training set was further subdivided into a learning set and a validation set using the same setup as described earlier. Furthermore, each of these models underwent training in ten different randomizations, and the best model weights were stored.

\subsection{Performance Metrics and Evaluation}
Out of the ten training randomizations for each model, the best weights were selected based on the criterion of achieving the maximum validation AUC-PRC. Both AUC-PRC and AUC-ROC are valuable metrics for assessing the prediction scores of a classifier. However, it was observed that our models consistently optimized AUC-ROC at higher values, rendering it an insufficiently discriminative metric for performance monitoring. What sets them apart is that AUC-PRC calculates the area under the Precision-Recall curve, incorporating precision and recall, whereas AUC-ROC employs the False Positive Rate (FPR) in place of precision. Unlike FPR both the precision and the recall are focused on the positive class (typically minority class in a imbalanced dataset) and are unconcerned with the true negatives (majority class in a imbalanced dataset). Therefore AUC-PRC is recommended for highly imbalanced datasets where AUC-ROC may provide an excessively optimistic view of the performance \cite{branco_survey_2016}.

After retaining the weights corresponding to the maximum validation AUC-PRC, each $LOPO_i$ model is subsequently employed to assess the performance on the test patient $P_i$. The performance was evaluated in terms of sensitivity, specificity, $F_1score$, positive predictive value(PPV) and the FAR/24h (false alarm rate per 24 hours). The AUC-ROC and AUC-PRC values of each LOPO model were also recorded. In addition to those patient-specific results, the overall results per each group were also evaluated. The best models trained using entire seizure-patient data from each group were used to evaluate FAR on the non-seizure patient data cohort. Furthermore, the overall FARs for each patient group were also recorded by aggregating the false alarm counts for \{$P_{s\_all}$\} and \{$P_{ns\_all}$\} and subsequently averaging them over the total recording duration of 5576 hours.

\begin{table}[]
\caption{THE LEAVE-ONE-PATIENT-OUT(LOPO) VALIDATION RESULTS FOR PATIENT GROUP PG1}
\label{table: PG1_results}
\resizebox{\columnwidth}{!}{%
\begin{tabular}{cccccccc}
\hline \hline
\begin{tabular}[c]{@{}c@{}}Patient\\  ID\end{tabular} & \begin{tabular}[c]{@{}c@{}}AUC\\ -ROC\end{tabular} & \begin{tabular}[c]{@{}c@{}}AUC\\ -PRC\end{tabular} & \begin{tabular}[c]{@{}c@{}}Sensitivity\\ (\%)\end{tabular} & \begin{tabular}[c]{@{}c@{}}Specificity\\ (\%)\end{tabular} & \begin{tabular}[c]{@{}c@{}}F1score\\ (\%)\end{tabular} & \begin{tabular}[c]{@{}c@{}}PPV\\ (\%)\end{tabular} & FAR/24h \\ \hline
$P_1$ & 1 & 1 & 100 & 100 & 100 & 100 & 0 \\
$P_2$ & 1 & 1 & 100 & 100 & 100 & 100 & 0 \\
$P_3$ & 1 & 1 & 100 & 100 & 100 & 100 & 0 \\
$P_4$ & 0.97 & 0.07 & 100 & 96.92 & 25 & 14.29 & 2.23 \\
$P_5$ & 1 & 1 & 100 & 100 & 100 & 100 & 0 \\
$P_6$ & 0.99 & 0.13 & 100 & 99.2 & 40 & 25 & 0.82 \\
$P_7$ & 0.97 & 0.25 & 100 & 97.37 & 66.67 & 50 & 0.28 \\
$P_8$ & 1 & 1 & 100 & 100 & 100 & 100 & 0 \\
$P_9$ & 0.99 & 0.13 & 100 & 98.83 & 40 & 25 & 0.77 \\
$P_{10}$ & 1 & 1 & 100 & 100 & 100 & 100 & 0 \\
$P_{11}$ & 1 & 1 & 100 & 100 & 100 & 100 & 0 \\
$P_{12}$ & 1 & 1 & 100 & 100 & 100 & 100 & 0 \\ \hline
Overall & 0.99 & 0.71 & \begin{tabular}[c]{@{}c@{}}100\\ (21/21)\end{tabular} & \begin{tabular}[c]{@{}c@{}}99.15\\ (1521/1534)\end{tabular} & 76.36 & 61.76 & 0.34 \\ \hline \hline
\end{tabular}%
}
\end{table}

\begin{table}[]
\caption{THE LEAVE-ONE-PATIENT-OUT(LOPO) VALIDATION RESULTS FOR PATIENT GROUP PG2}
\label{table: PG2_results}
\resizebox{\columnwidth}{!}{%
\begin{tabular}{cccccccc}
\hline \hline
\begin{tabular}[c]{@{}c@{}}Patient\\ ID\end{tabular} & \begin{tabular}[c]{@{}c@{}}AUC\\ -ROC\end{tabular} & \begin{tabular}[c]{@{}c@{}}AUC\\ -PRC\end{tabular} & \begin{tabular}[c]{@{}c@{}}Sensitivity\\ (\%)\end{tabular} & \begin{tabular}[c]{@{}c@{}}Specificity\\ (\%)\end{tabular} & \begin{tabular}[c]{@{}c@{}}F1score\\ (\%)\end{tabular} & \begin{tabular}[c]{@{}c@{}}PPV\\ (\%)\end{tabular} & FAR/24h \\ \hline 
$P_1$ & 1 & 1 & 100 & 100 & 100 & 100 & 0 \\
$P_2$ & 0.99 & 0.71 & 100 & 97.65 & 66.67 & 50 & 0.59 \\
$P_3$ & 1 & 1 & 100 & 100 & 100 & 100 & 0 \\
$P_4$ & 0.98 & 0.1 & 100 & 97.95 & 33.33 & 20 & 1.49 \\
$P_5$ & 1 & 1 & 100 & 100 & 100 & 100 & 0 \\
$P_6$ & 1 & 1 & 100 & 100 & 100 & 100 & 0 \\
$P_7$ & 0.97 & 0.25 & 100 & 97.37 & 66.67 & 50 & 0.28 \\
$P_8$ & 1 & 1 & 100 & 100 & 100 & 100 & 0 \\
$P_9$ & 0.99 & 0.17 & 100 & 99.22 & 50 & 33.33 & 0.51 \\
$P_{10}$ & 1 & 1 & 100 & 100 & 100 & 100 & 0 \\
$P_{11}$ & 1 & 1 & 100 & 100 & 100 & 100 & 0 \\
$P_{12}$ & 0.68 & 0.68 & 66.67 & 100 & 80 & 100 & 0 \\
$P_{19}$ & 1 & 1 & 100 & 100 & 100 & 100 & 0 \\
$P_{20}$ & 1 & 1 & 100 & 100 & 100 & 100 & 0 \\ \hline 
Overall & 0.97 & 0.78 & \begin{tabular}[c]{@{}c@{}}96.15\\ (25/26)\end{tabular} & \begin{tabular}[c]{@{}c@{}}99.44\\ (1593/1602)\end{tabular} & 83.33 & 73.53 & 0.2\\ \hline \hline
\end{tabular}%
}
\end{table}

\begin{table}[]
\caption{THE LEAVE-ONE-PATIENT-OUT(LOPO) VALIDATION RESULTS FOR PATIENT GROUP PG3}
\label{table: PG3_results}
\resizebox{\columnwidth}{!}{%
\begin{tabular}{cccccccc}
\hline \hline
\begin{tabular}[c]{@{}c@{}}Patient\\ ID\end{tabular} & \begin{tabular}[c]{@{}c@{}}AUC\\ -ROC\end{tabular} & \begin{tabular}[c]{@{}c@{}}AUC\\ -PRC\end{tabular} & \begin{tabular}[c]{@{}c@{}}Sensitivity\\ (\%)\end{tabular} & \begin{tabular}[c]{@{}c@{}}Specificity\\ (\%)\end{tabular} & \begin{tabular}[c]{@{}c@{}}F1score\\ (\%)\end{tabular} & \begin{tabular}[c]{@{}c@{}}PPV\\ (\%)\end{tabular} & FAR/24h \\ \hline 
$P_1$ & 1 & 1 & 100 & 100 & 100 & 100 & 0 \\
$P_2$ & 1 & 1 & 100 & 100 & 100 & 100 & 0 \\
$P_3$ & 1 & 1 & 100 & 100 & 100 & 100 & 0 \\
$P_4$ & 0.87 & 0.02 & 0 & 100 & 0 & 0 & 0 \\
$P_5$ & 1 & 1 & 100 & 100 & 100 & 100 & 0 \\
$P_6$ & 1 & 1 & 100 & 100 & 100 & 100 & 0 \\
$P_7$ & 0.92 & 0.13 & 100 & 92.11 & 40 & 25 & 0.83 \\
$P_8$ & 1 & 1 & 100 & 100 & 100 & 100 & 0 \\
$P_9$ & 0.99 & 0.13 & 100 & 98.83 & 40 & 25 & 0.77 \\
$P_{10}$ & 1 & 1 & 100 & 100 & 100 & 100 & 0 \\
$P_{11}$ & 1 & 1 & 100 & 100 & 100 & 100 & 0 \\
$P_{12}$ & 1 & 1 & 100 & 100 & 100 & 100 & 0 \\
$P_{13}$ & 1 & 1 & 100 & 100 & 100 & 100 & 0 \\
$P_{14}$ & 0.95 & 0.34 & 100 & 86.14 & 41.67 & 26.32 & 5.54 \\
$P_{15}$ & 0.77 & 0.01 & 0 & 100 & 0 & 0 & 0 \\
$P_{16}$ & 0.68 & 0.02 & 0 & 100 & 0 & 0 & 0 \\
$P_{17}$ & 0.98 & 0.78 & 100 & 95 & 78.26 & 64.29 & 2.64 \\
$P_{18}$ & 0.98 & 0.05 & 0 & 100 & 0 & 0 & 0 \\ \hline
Overall & 0.95 & 0.64 & \begin{tabular}[c]{@{}c@{}}90.24\\ (37/41)\end{tabular} & \begin{tabular}[c]{@{}c@{}}99.16\\ (2953/2978)\end{tabular} & 71.84 & 59.68 & 0.54\\ \hline \hline
\end{tabular}%
}
\end{table}

\begin{table}[]
\caption{THE LEAVE-ONE-PATIENT-OUT(LOPO) VALIDATION RESULTS FOR PATIENT GROUP PG4}
\label{table: PG4_results}
\resizebox{\columnwidth}{!}{%
\begin{tabular}{cccccccc}
\hline \hline
\begin{tabular}[c]{@{}c@{}}Patient\\ ID\end{tabular} & \begin{tabular}[c]{@{}c@{}}AUC\\ -ROC\end{tabular} & \begin{tabular}[c]{@{}c@{}}AUC\\ -PRC\end{tabular} & \begin{tabular}[c]{@{}c@{}}Sensitivity\\ (\%)\end{tabular} & \begin{tabular}[c]{@{}c@{}}Specificity\\ (\%)\end{tabular} & \begin{tabular}[c]{@{}c@{}}F1score\\ (\%)\end{tabular} & \begin{tabular}[c]{@{}c@{}}PPV\\ (\%)\end{tabular} & FAR/24h \\ \hline 
$P_1$ & 1 & 1 & 100 & 100 & 100 & 100 & 0 \\
$P_2$ & 0.99 & 0.76 & 100 & 98.81 & 85.71 & 75 & 0.29 \\
$P_3$ & 1 & 1 & 100 & 100 & 100 & 100 & 0 \\
$P_4$ & 0.88 & 0.02 & 0 & 100 & 0 & 0 & 0 \\
$P_5$ & 1 & 1 & 100 & 100 & 100 & 100 & 0 \\
$P_6$ & 1 & 1 & 100 & 100 & 100 & 100 & 0 \\
$P_7$ & 0.95 & 0.17 & 100 & 94.74 & 50 & 33.33 & 0.56 \\
$P_8$ & 1 & 1 & 100 & 100 & 100 & 100 & 0 \\
$P_9$ & 0.99 & 0.17 & 100 & 99.22 & 50 & 33.33 & 0.51 \\
$P_{10}$ & 1 & 1 & 100 & 100 & 100 & 100 & 0 \\
$P_{11}$ & 1 & 1 & 100 & 100 & 100 & 100 & 0 \\
$P_{12}$ & 1 & 0.77 & 66.67 & 100 & 80 & 100 & 0 \\
$P_{13}$ & 1 & 1 & 100 & 100 & 100 & 100 & 0 \\
$P_{14}$ & 0.96 & 0.37 & 100 & 92.08 & 55.56 & 38.46 & 3.17 \\
$P_{15}$ & 0.98 & 0.1 & 100 & 98.33 & 33.33 & 20 & 1.89 \\
$P_{16}$ & 0.93 & 0.06 & 100 & 92.93 & 22.22 & 12.5 & 2.39 \\
$P_{17}$ & 0.99 & 0.9 & 100 & 94 & 75 & 60 & 3.17 \\
$P_{18}$ & 0.99 & 0.08 & 100 & 98.85 & 28.57 & 16.67 & 1.92 \\
$P_{19}$ & 1 & 1 & 100 & 100 & 100 & 100 & 0 \\
$P_{20}$ & 1 & 1 & 100 & 100 & 100 & 100 & 0 \\ \hline
Overall & 0.98 & 0.67 & \begin{tabular}[c]{@{}c@{}}95.65\\ (44/46)\end{tabular} & \begin{tabular}[c]{@{}c@{}}98.85\\ (3011/3046)\end{tabular} & 70.4 & 55.7 & 0.69\\ \hline \hline
\end{tabular}%
}
\end{table}

\section{Results}
All patient-specific and overall results are presented in Tables \ref{table: PG1_results} to \ref{table: PG4_results}. The results for patient group PG1 (Table \ref{table: PG1_results}) show that our DNN correctly classified all GTCS events (100\%) and 1521 (99.25\%) out of 1534 non-seizure events, with a mean FAR of 0.34/24h (13 false alarms from 12 patients). The overall $F_1$score and PPV were 76.36\% and 61.76\%, respectively. Notably, the FAR for the 67 patients without any GTCS was 0, resulting in an overall FAR of 0.06/24h for 79 patients over the total recording period of 5576 hours.

As shown in Table \ref{table: PG2_results}, the DNN detected 25 (96.15\%) out of 26 seizure events and 1593 (99.44\%) out of 1602 non-seizure events, achieving an $F_1$score of 83.33\% and a PPV of 73.53\%. One FIAS event went undetected in the case of patient $P_{12}$. The mean FAR, evaluated on 14 patients, was 0.2/24h (9 false alarms), while it was 0.01/24h for the remaining 65 patients without any GTCS or FIAS (2 false alarms). This resulted in an overall FAR of 0.05/24h (11 false alarms for 79 patients).

Table \ref{table: PG3_results} presents the results for patient group PG3. The DNN correctly detected 37 (90.24\%) out of 41 seizure events, with the four missed seizures (1 GTCS and 3 PNES) corresponding to patients $P_{4}$, $P_{15}$, $P_{16}$, and $P_{18}$. Additionally, it accurately classified 2953 (99.16\%) out of 2978 non-seizure events. The overall $F_1$ score and PPV were 71.84\% and 59.68\%, respectively. The mean FAR for the 18 patients was 0.54/24h, with 25 false alarms. During validation on 61 patients without GTCS or PNES, there were 7 false alarms, resulting in a FAR of 0.04/24h. The overall FAR for all 79 patients was 0.14/24h, with 32 false alarms.

The results for patient group PG4 (Table \ref{table: PG4_results}) indicate that the DNN could detect 44 (95.65\%) out of 46 total seizure events, encompassing all three seizure types: GTCS, FIAS, and PNES. Additionally, it correctly identified 3011 (98.85\%) out of 3046 non-seizure events, with a mean FAR of 0.69/24h across 20 patients (35 false alarms). The resulting $F_1$score and PPV were 70.4\% and 55.7\%, respectively. Notably, two events (one GTCS and one FIAS) went undetected in patients $P_{4}$ and $P_{12}$. When evaluated on 59 patients without any type of seizure, there was one false alarm, resulting in an overall FAR of 0.15/24h over the 5576h recording period for 79 patients (36 false alarms).

\section{Discussion}
In this research paper, our primary objective is to employ wearable accelerometry for detecting convulsive epileptic seizures. Distinguishing our approach from the majority of existing studies within the literature, which often involve the use of multiple modalities or sensors[], our methodology exclusively relies on a solitary accelerometer sensor for data acquisition. Notably, this system can effectively identify various seizure types, namely, 1) Generalized tonic-clonic seizure, 2) psychogenic non-epileptic seizures, and 3) focal impaired awareness(previously complex partial) seizures in contrast to prevalent research which predominantly focuses on Generalized Tonic-Clonic Seizures (GTCS) or other isolated seizure types[]. Based on our dataset, it is evident that the co-occurrence of multiple seizure types within a single patient is possible (1 patient with GTCS + FIAS, and 1 patient with GTCS + PNES). Furthermore, the occurrence of CPS and PNES alone is also noteworthy, aside from GTCS (60\% of 20 patients had GTCS, 30\% had PNES and 10\% had FIAS(Table \ref{table:overview_of_patients})). Therefore, it is important for seizure detection systems to be capable of detecting multiple seizure types, as this better reflects their results within the general epileptic population. Our model, when applied to the patient group PG4 (comprising all seizure types), achieved an AUC of 0.98 and an AUC-PRC of 0.67, demonstrating the feasibility of such a system employing wearable ACM. Additionally, our method is capable of detecting seizures with short durations ($\geq$ 10 seconds). In combination with its ability to detect multiple seizure types, this contributes to the accurate recording of seizure frequencies in seizure diaries. This, in turn, assists clinicians in better managing treatments. This section presents a comprehensive performance evaluation and a comparison of results with the current literature.

\subsection{Performance of the Proposed Method}
For the patient group PG1 (consisting only of GTCS), all GTCS events were successfully detected. Eight out of the 12 patients achieved an AUC of 1.0 and an AUC-PRC of 1.0 with no false alarms. Surprisingly, among these eight patients, two had multiple seizure types ($P_2$ with 2 GTCS and 1 PNES and $P_{12}$ with 2 GTCS and 1 FIAS, as shown in Table \ref{table:overview_of_patients}). This indicates that the sensitivity and FAR of our seizure detection model for GTCS remain intact even in the presence of multiple types of seizures. Notably, $P_4$ had the minimum AUC-PRC (Table \ref{table: PG1_results}) and the highest FAR which accounts for nearly half of the false alarms among 20 patients. 
Based on visual inspections, it was observed that the GTCS event in patient $P_4$ exhibited less pronounced characteristics of having sharp, transient, and concentrated spikes, which are typically observed in other GTCS signals. Instead, it consisted with more dispersed spikes. This observation may be attributed to the presence of the clonic phase, which was not clearly discernible in the seizure manifestation. Therefore, the missing seizure can be identified as a comparatively less violent seizure, lacking the pronounced high-frequency twitches or clonic rhythmical jerks, typically associated with GTCS\cite{seizureTypeClassification}.

The patient group PG2, (considering both GTCS and FIAS), missed only one FIAS seizure in the case of patient $P_{12}$. Patient $P_{12}$ exhibited  multiple seizure types, and while all instances of GTCS were detected, this particular FIAS seizure went undetected. It is worth noting that FIAS seizures are frequently underreported due to their variable presentation; the presence of rhythmic jerking in a single limb is optional, and some patients may exhibit no discernible symptoms at all\cite{LancetDiagnosticChallengesEpilepsy, seizureTypeClassification}. This fact is further substantiated by the visual inspection of ACM signals, where GTCS events are typically characterized by concentrated sudden bursts of activity, while FIAS signals lack this nature and exhibit similarities to non-seizure events. 

However, the seizure events of patients $P_{19}$ and $P_{20}$, who  exclusively experienced FIAS (1 for patient $P_{19}$ and 3 for patient $P_{20}$), were impeccably classified, resulting in zero false alarms. This outcome is particularly surprising given the limited exposure to FIAS events during the model's training phase, which was further exacerbated in the Leave-One-Patient-Out (LOPO) setting. For instance, in the LOPO model for patient $P_{20}$, only 2 FIAS training samples were available out of the total 5 FIAS in the training dataset. This suggests that the model's ability to detect FIAS benefited from the inherent structural knowledge it gained from learning GTCS patterns.

From the patient group PG3, which includes both GTCS and PNES, a total of 4 seizures went undetected, with 3 of them being PNES.  Test results for ten patients among the fourteen, achieved an AUCPRC of 1.0 and an AUCROC of 1.0, with no false alarms recorded. Notably, this excellent performance extends to the test results for the patient $P_2$, who experienced both GTCS and PNES (refer to Table \ref{table: PG3_results}). In the case of the test results for the 6 patients who exclusively had PNES events, the majority exhibited relatively lower AUCPRC values, contributing to 75\% of the total false alarms. The overall PNES detection rate reached 85\%, while the GTCS detection rate was notably higher at 95\%.

However, the models for patients $P_{14}$ and $P_{17}$, who experienced 5 PNES and 9 PNES respectively, were able to detect all PNES events, even though during LOPO validation, nearly half of the PNES events were available for training. This demonstrates that models trained on ACM sensor data, which captures behavioural changes, can leverage the knowledge gained from GTCS to detect PNES as well. However, if we were to use EEG sensor modality instead, which captures electrical brain activity, the applicability of knowledge from GTCS for identifying PNES diminishes. This is due to the fact that, unlike in epileptic seizures, PNES are not associated with abnormal electrical activity in the brain, resulting in EEG recordings during seizures remaining normal without displaying ictal electrical changes\cite{brown_psychological_2016, lafrance_jr_minimum_2013,gedzelman_long-term_2014,ahmadi_eeg-based_2020}. This fundamental distinction sets PNES apart from GTCS in terms of EEG characteristics.
  
A total of 2 seizures, comprising 1 GTCS and 1 FIAS, went undetected in patient group PG4, which encompasses all seizure types (see Table \ref{table: PG4_results}). It's worth noting that the FIAS event that went undetected was the same event from patient $P_{12}$ that was not detected in PG2 as well. However, this time, the test results for patient $P_{12}$ showed an improvement in both the AUCPRC and AUCROC compared to PG3. This observation demonstrates that the inclusion of PNES for training contributed to enhancing the model's performance in distinguishing FIAS from normal events.

Similar to PG3, the test results for patients who exclusively had PNES in PG4 achieved relatively lower test AUCPRC and AUCROC values, contributing to 85\% of the total false alarms. However, this time, the AUCROCs showed a slight improvement compared to PG3. This suggests that including both FIAS and PNES in the training data can mutually benefit each other in producing optimal models to detect these seizure types from non-seizure events. Furthermore, the GTCS event that went undetected in PG4 was from the same patient $P_4$, who previously contributed to missing GTCS events or generating the highest FAR in previous groups. 

While these results are impressive, it's essential to acknowledge that firm inferences regarding FIAS cannot be drawn due to the limited representation of FIAS in this dataset (only 5 FIAS events). However, the inclusion of FIAS data was crucial because it represented an important patient group necessary to model a more generalized epileptic cohort. This is of particular significance because FIASs are the most common type of known seizure in adults with epilepsy.\cite{mostCommonType,commonSeizuresSurvay, keranen_distribution_1988}.

\begin{table}[]
\caption{Result Comparison Of The Proposed Method With The Baseline Method}
\label{Table:comp_to_baseline}
\resizebox{\columnwidth}{!}{%
\begin{tabular}{lllllll}
\hline \hline
{\color[HTML]{000000} \textbf{patient group}} & {\color[HTML]{000000} \textbf{method}} & {\color[HTML]{000000} \textbf{sensitivity}} & {\color[HTML]{000000} \textbf{specificity}} & {\color[HTML]{000000} \textbf{F1score}} & {\color[HTML]{000000} \textbf{PPV}} & {\color[HTML]{000000} \textbf{FAR/24h}} \\ \hline 
GTCS & {\color[HTML]{2F75B5} baseline} & {\color[HTML]{2F75B5} 95.23} & {\color[HTML]{2F75B5} 98.04} & {\color[HTML]{2F75B5} 56.33} & {\color[HTML]{2F75B5} 40} & {\color[HTML]{2F75B5} 0.72} \\
 & {\color[HTML]{C65911} proposed} & {\color[HTML]{C65911} 100} & {\color[HTML]{C65911} 99.15} & {\color[HTML]{C65911} 76.36} & {\color[HTML]{C65911} 61.76} & {\color[HTML]{C65911} 0.34} \\ \hline
GTCS+CPS & {\color[HTML]{305496} baseline} & {\color[HTML]{305496} 88.26} & {\color[HTML]{305496} 96.75} & {\color[HTML]{305496} 45.54} & {\color[HTML]{305496} 30.66} & {\color[HTML]{305496} 1.24} \\
 & {\color[HTML]{C65911} proposed} & {\color[HTML]{C65911} 96.15} & {\color[HTML]{C65911} 99.44} & {\color[HTML]{C65911} 83.33} & {\color[HTML]{C65911} 73.53} & {\color[HTML]{C65911} 0.2} \\ \hline
GTCS+PNES & {\color[HTML]{305496} baseline} & {\color[HTML]{305496} 87.8} & {\color[HTML]{305496} 97.12} & {\color[HTML]{305496} 44.44} & {\color[HTML]{305496} 29.75} & {\color[HTML]{305496} 1.56} \\
 & {\color[HTML]{C65911} proposed} & {\color[HTML]{C65911} 90.24} & {\color[HTML]{C65911} 99.16} & {\color[HTML]{C65911} 71.84} & {\color[HTML]{C65911} 59.68} & {\color[HTML]{C65911} 0.54} \\ \hline
ALL & {\color[HTML]{305496} baseline} & {\color[HTML]{305496} 86.95} & {\color[HTML]{305496} 96.16} & {\color[HTML]{305496} 39.6} & {\color[HTML]{305496} 25.64} & {\color[HTML]{305496} 2.25} \\
 & {\color[HTML]{C65911} proposed} & {\color[HTML]{C65911} 95.65} & {\color[HTML]{C65911} 98.85} & {\color[HTML]{C65911} 70.4} & {\color[HTML]{C65911} 55.7} & {\color[HTML]{C65911} 0.69} \\ \hline \hline
\end{tabular}%
}
\end{table}

\begin{table*}[t]
\centering
\caption{Comparison of the proposed study with seizure detection systems based on a single sensor.}
\label{Table:single_sensor_comp}

\scriptsize
\setlength{\tabcolsep}{2.5pt}
\renewcommand{\arraystretch}{1.15}

\resizebox{\textwidth}{!}{%
\begin{tabular}{@{}lccccccccccccccc@{}}
\toprule

\textbf{Study} &
\textbf{baseline\cite{baseline}} &
\textbf{Beniczky \cite{beniczky}} &
\textbf{Lockman \cite{lockman}} &
\textbf{Kramer\cite{kramer}} &
\textbf{Conradsen\cite{conradsen}} &
\textbf{Szab\'o\cite{szabo}} &
\textbf{Agrahri\cite{agrahriAnushman2022}} &
\textbf{Motahar\cite{motaharErina2021}} &
\textbf{Beniczky\cite{beniczky2018}} &
\textbf{Jeppesen\cite{jeppesen_seizure_2019}} &
\textbf{CNN} &
\begin{tabular}[c]{@{}c@{}}\textbf{LSTMFCN}\\\textbf{tsai}\end{tabular} &
\begin{tabular}[c]{@{}c@{}}\textbf{ResNet}\\\textbf{tsai}\end{tabular} &
\begin{tabular}[c]{@{}c@{}}\textbf{CNN}\\\textbf{Keras (all)}\end{tabular} &
\begin{tabular}[c]{@{}c@{}}\textbf{ResNet}\\\textbf{tsai (all)}\end{tabular} \\
\midrule

\multicolumn{16}{@{}l}{\textit{\textbf{Fixed parameters}}} \\[2pt]

\textbf{Sensor} &
1 ACM & 1 ACM & 1 ACM & 1 ACM &
1 sEMG & 1 sEMG & 1 ACM & 1 ACM &
1 sEMG & 1 ECG (HRV) &
1 ACM & 1 ACM & 1 ACM & 1 ACM & 1 ACM \\

\addlinespace[3pt]
\multicolumn{16}{@{}l}{\textit{\textbf{Variable parameters}}} \\[2pt]

\textbf{Seizure duration} &
$\geq 10$ & -- & -- & $\geq 20$ &
$\geq 50$ & -- & -- & 1 s &
-- & $\geq 20$ &
$\geq 10$ & $\geq 10$ & $\geq 10$ & $\geq 10$ & $\geq 10$ \\

\textbf{Total patients} &
79 & 73 & 40 & 31 & 60 & 33 & 79 & 1 &
71 & 19 & 79 & 79 & 79 & 79 & 79 \\

\textbf{Seizure patients} &
12 & 20 & 6 & 15 & 11 & 11 & 35 & 1 &
20 & 11 & 12 & 12 & 12 & 20 & 20 \\

\textbf{Total seizures} &
21 & 39 & 8 & 22 & 22 & 21 & 46 & 125 &
32 & 23 & 21 & 21 & 21 & 46 & 46 \\

\textbf{Seizure type} &
GTCS & GTCS & GTCS & C, GTCS &
GTCS & GTCS & -- &
\begin{tabular}[c]{@{}c@{}}Synthetic\\seizures\end{tabular} &
GTCS &
\begin{tabular}[c]{@{}c@{}}TCS,\\Focal\end{tabular} &
GTCS & GTCS & GTCS &
\begin{tabular}[c]{@{}c@{}}GTCS, PNES,\\CPS\end{tabular} &
\begin{tabular}[c]{@{}c@{}}GTCS, PNES,\\CPS\end{tabular} \\

\textbf{Record duration} &
5576 h & 4878 h & -- & 1692 h &
776 h & 1399 h & -- & 184 min &
3735.5 h & 1434 h &
5556.5 h & 5556.5 h & 5556.5 h & 5556.5 h & 5556.5 h \\

\midrule

\textbf{Sensitivity} &
\begin{tabular}[c]{@{}c@{}}95.23\%\\(20/21)\end{tabular} &
\begin{tabular}[c]{@{}c@{}}90\%\\(35/39)\end{tabular} &
\begin{tabular}[c]{@{}c@{}}88\%\\(7/8)\end{tabular} &
\begin{tabular}[c]{@{}c@{}}91\%\\(20/22)\end{tabular} &
\begin{tabular}[c]{@{}c@{}}100\%\\(22/22)\end{tabular} &
\begin{tabular}[c]{@{}c@{}}95\%\\(20/21)\end{tabular} &
\begin{tabular}[c]{@{}c@{}}95.65\%\\(44/46)\end{tabular} &
\begin{tabular}[c]{@{}c@{}}98.4\%\\(123/125)\end{tabular} &
\begin{tabular}[c]{@{}c@{}}93.8\%\\(30/32)\end{tabular} &
87\% &
\begin{tabular}[c]{@{}c@{}}100\%\\(21/21)\end{tabular} &
\begin{tabular}[c]{@{}c@{}}95.23\%\\(20/21)\end{tabular} &
\begin{tabular}[c]{@{}c@{}}100\%\\(21/21)\end{tabular} &
\begin{tabular}[c]{@{}c@{}}97.83\\(45/46)\end{tabular} &
\begin{tabular}[c]{@{}c@{}}95.65\%\\(44/46)\end{tabular} \\

\textbf{FAR} &
0.64/24 h & 0.2/24 h & -- & 0.11/24 h &
0.96/24 h & 0.02/24 h & 14.80\% & 0\% &
0.67/24 h & 0.9/24 h &
0.29/24 h & 0.02/24 h & 0.06/24 h & 0.58/24 h & 0.15/24 h \\

\bottomrule
\end{tabular}%
}

\vspace{2pt}
\begin{minipage}{\textwidth}
\scriptsize
\textit{Note:} ``--'' indicates that the corresponding value was not reported.
ACM: accelerometry; sEMG: surface electromyography; ECG: electrocardiography;
HRV: heart-rate variability; FAR: false alarm rate.
\end{minipage}

\end{table*}

\subsection{Comparison with Existing Litterateur}
First, the results are compared with the previous work \cite{baseline}, which uses the exact same dataset as ours and is considered the baseline for our study. Table \ref{Table:comp_to_baseline} summarizes the results for the proposed method and the baseline method \cite{baseline} for all patient groups. All sensitivities, PPV, F1Score and FAR values demonstrate outstanding improvements across all categories when compared to the baseline. Notably, sensitivity improvements are most pronounced in PG2 and PG4, where they improve by nearly 9\%. On the other hand, the remarkable reduction in FAR stands as a significant achievement across patient groups: in PG1, it was halved (reduced by 50\%), in PG2, it was notably reduced to one-sixth (16\%), and in PG3 and PG4, it was  reduced to approximately one-third (34\% and 30\% respectively) of their baseline values. This demonstrates the superior potential of the ResNet deep learning classifier\cite{he_deep_2015,resnet_TSC} in seizure detection compared to the SVDD method, which relies solely on hand-crafted features.

Comparison of the proposed model with existing studies are presented below, following the same style proposed by the baseline paper.
The main contributing factors that affect the performance of seizure detection systems, namely, 1) seizure type, 2) the minimum seizure duration, 3) the number of sensors, and 4) multi/single modality, were taken one by one and kept fixed across the studies to enable a justifiable comparison. 

Table \ref{Table:single_sensor_comp} compares the implemented deep learning models in this study with other seizure detection systems when number of sensors fixed to a single sensor. The implemented ResNet model for GTCS seizures only (PG1), achieved the better performance in-terms of both the sensitivity and FAR compared to Kusmakar \textit{et al}. \cite{baseline}, Beniczky \textit{et al}. \cite{beniczky}, Lockman \textit{et al}. \cite{lockman}, Kramer \textit{et al}. \cite{kramer}, Beniczky \textit{et al}. 2018 \cite{beniczky2018} and Jeppesen \textit{et al}. \cite{jeppesen_seizure_2019}. 
Interestingly, our system could detect seizures of shorter duration than the system proposed by Kramer \textit{et al}. \cite{kramer}, but our FAR is significantly better than theirs, even though it was expected to increase when reducing the minimum seizure duration\cite{baseline}.
In another study that employed one sEMG sensor, Conradsen \textit{et al}. \cite{conradsen} achieved the same seizure detection sensitivity as ours, which is 100\%, but with a considerably higher FAR of 0.96/24h. Furthermore, their minimum seizure duration is longer (50s), and the assurance of achieving a similar high sensitivity in shorter-duration seizures is somewhat uncertain. 
In a distinct study conducted by Szab\'o \textit{et al.} \cite{szabo}, they achieved a lower FAR compared to our system. Nevertheless, their methodology also yielded a lower sensitivity value when compared with our system's performance. Furthermore, it is noteworthy that the total recording duration in their study accounted for approximately one-fourth of the total recording duration employed in our research, a factor that may have contributed to the observed lower FAR. 
Motahar \textit{et al.} \cite{motaharErina2021} achieved 98.40\% of seizure detection sensitivity which is slightly less compared to ours, and with zero false alarms. However, there are concerns regarding the generalizability of their findings to real patient data, as their model was developed and evaluated using synthetic seizure data gathered via simulated seizure-like movements. Meanwhile Agrahri \textit{et al.} \cite{agrahriAnushman2022} reported the performance of their seizure detection systems with a sensitivity of 95.65\% and a FAR of 14.80\%. In terms of sensitivity, our system performs better compared to them. However, a direct comparison of FAR values between the two studies is not feasible due to the absence of total record duration information in \cite{agrahriAnushman2022}.  Nevertheless, their reported count of 451 false alarms stands significantly higher in contrast to the 13 false alarms generated by our model.

\section{Conclusion}
The application of automated seizure detection using non-invasive wearable devices holds the potential to significantly reduce the morbidity and mortality associated with seizures, providing objective seizure documentation and improving the quality of life for individuals with epilepsy, particularly if these systems maintain high seizure detection sensitivities and low FARs.
Our study proposed a remote seizure detection system employing a single wrist-worn accelerometer sensor, capable of detecting three major seizure types with durations $\geq 10 $ s. When validated on accelerometer data from patients undergoing VEM, our system successfully detected 44 out of 46 convulsive seizures, achieving a sensitivity of 95.65\% with an overall FAR of 0.15/24 h. Importantly, this proposed system is a generalized model, making it applicable to any patient without the need for personalized data collection.
Utilizing deep neural networks of sufficient depth empowers us to grasp the complexity within the data, thus enabling us to achieve a very lower FAR while retaining a high level of sensitivity.
The pre-processing steps and weighted loss function effectively address the class imbalance problem, facilitating the modeling of the problem as a supervised learning method. Furthermore, the comparative analysis with existing literature demonstrates the feasibility and potential of our proposed system for non-invasive automated seizure detection.
By utilizing the time-series ResNet, our system autonomously extracts optimal features for multiple seizure types without the need for manual feature engineering. This adaptability positions our system to easily accommodate the detection of other seizure types as relevant training data becomes available in the future. 
Overall, our study contributes to the ongoing efforts to enhance the quality of care and management for individuals living with epilepsy through effective non-invasive seizure detection method.

\section*{Acknowledgment}
The authors would like to thank the patients of Melbourne Brain Center, Royal Melbourne Hospital, who contributed their data for the study. 

\bibliographystyle{plain}
\bibliography{references12}

@article{beniczky_automated_2021,
	title = {Automated seizure detection using wearable devices: {A} clinical practice guideline of the {International} {League} {Against} {Epilepsy} and the {International} {Federation} of {Clinical} {Neurophysiology}},
	volume = {132},
	issn = {1388-2457},
	shorttitle = {Automated seizure detection using wearable devices},
	url = {https://www.sciencedirect.com/science/article/pii/S1388245720306143},
	doi = {10.1016/j.clinph.2020.12.009},
	number = {5},
	urldate = {2023-11-14},
	journal = {Clinical Neurophysiology},
	author = {Beniczky, Sándor and Wiebe, Samuel and Jeppesen, Jesper and Tatum, William O. and Brazdil, Milan and Wang, Yuping and Herman, Susan T. and Ryvlin, Philippe},
	month = may,
	year = {2021},
	pages = {1173--1184},
}

@article{simblett_patients_2020,
	title = {Patients' experience of wearing multimodal sensor devices intended to detect epileptic seizures: {A} qualitative analysis},
	volume = {102},
	issn = {1525-5069},
	shorttitle = {Patients' experience of wearing multimodal sensor devices intended to detect epileptic seizures},
	doi = {10.1016/j.yebeh.2019.106717},
	language = {eng},
	journal = {Epilepsy \& Behavior: E\&B},
	author = {Simblett, Sara Katherine and Biondi, Andrea and Bruno, Elisa and Ballard, Dominic and Stoneman, Amanda and Lees, Simon and Richardson, Mark P. and Wykes, Til and {RADAR-CNS consortium}},
	month = jan,
	year = {2020},
	pmid = {31785481},
	pages = {106717},
}

@inproceedings{wang_time_2017,
	title = {Time series classification from scratch with deep neural networks: {A} strong baseline},
	shorttitle = {Time series classification from scratch with deep neural networks},
	url = {https://ieeexplore.ieee.org/document/7966039},
	doi = {10.1109/IJCNN.2017.7966039},
	urldate = {2023-10-21},
	booktitle = {2017 {International} {Joint} {Conference} on {Neural} {Networks} ({IJCNN})},
	author = {Wang, Zhiguang and Yan, Weizhong and Oates, Tim},
	month = may,
	year = {2017},
	note = {ISSN: 2161-4407},
	pages = {1578--1585},
}

@article{bayly_time-frequency_2013,
	title = {Time-frequency mapping of the rhythmic limb movements distinguishes convulsive epileptic from psychogenic nonepileptic seizures},
	volume = {54},
	issn = {1528-1167},
	url = {https://onlinelibrary.wiley.com/doi/abs/10.1111/epi.12207},
	doi = {10.1111/epi.12207},
	language = {en},
	number = {8},
	urldate = {2023-10-01},
	journal = {Epilepsia},
	author = {Bayly, Jade and Carino, John and Petrovski, Slavé and Smit, Michelle and Fernando, Dilini A. and Vinton, Anita and Yan, Bernard and Gubbi, Jayavardhana R. and Palaniswami, Marimuthu S. and O'Brien, Terence J.},
	year = {2013},
	note = {\_eprint: https://onlinelibrary.wiley.com/doi/pdf/10.1111/epi.12207},
	pages = {1402--1408},
}

@article{gubbi_automatic_2016,
	title = {Automatic {Detection} and {Classification} of {Convulsive} {Psychogenic} {Nonepileptic} {Seizures} {Using} a {Wearable} {Device}},
	volume = {20},
	issn = {2168-2208},
	url = {https://ieeexplore.ieee.org/document/7126915},
	doi = {10.1109/JBHI.2015.2446539},
	number = {4},
	urldate = {2023-10-01},
	journal = {IEEE Journal of Biomedical and Health Informatics},
	author = {Gubbi, Jayavardhana and Kusmakar, Shitanshu and Rao, Aravinda S. and Yan, Bernard and O’Brien, Terence and Palaniswami, Marimuthu},
	month = jul,
	year = {2016},
	note = {Conference Name: IEEE Journal of Biomedical and Health Informatics},
	pages = {1061--1072},
}

@article{kingma_adam_2014,
	title = {Adam: {A} {Method} for {Stochastic} {Optimization}},
	shorttitle = {Adam},
	journal = {International Conference on Learning Representations},
	author = {Kingma, Diederik and Ba, Jimmy},
	month = dec,
	year = {2014},
}

@article{zeiler_adadelta_2012,
	title = {{ADADELTA}: {An} adaptive learning rate method},
	volume = {1212},
	shorttitle = {{ADADELTA}},
	author = {Zeiler, Matthew},
	month = dec,
	year = {2012},
}

@article{duchi_adaptive_nodate,
	title = {Adaptive {Subgradient} {Methods} for {Online} {Learning} and {Stochastic} {Optimization}},
	language = {en},
	author = {Duchi, John and Hazan, Elad and Singer, Yoram},
}

@inproceedings{smith_cyclical_2017,
	title = {Cyclical {Learning} {Rates} for {Training} {Neural} {Networks}},
	doi = {10.1109/WACV.2017.58},
	author = {Smith, Leslie},
	month = mar,
	year = {2017},
	pages = {464--472},
}

@article{smith_super-convergence_2018,
	title = {Super-{Convergence}: {Very} {Fast} {Training} of {Residual} {Networks} {Using} {Large} {Learning} {Rates}},
	shorttitle = {Super-{Convergence}},
	url = {https://openreview.net/forum?id=H1A5ztj3b},
	language = {en},
	urldate = {2023-09-24},
	author = {Smith, Leslie N. and Topin, Nicholay},
	month = feb,
	year = {2018},
}

@article{branco_survey_2016,
	title = {A {Survey} of {Predictive} {Modeling} on {Imbalanced} {Domains}},
	volume = {49},
	issn = {0360-0300},
	url = {https://dl.acm.org/doi/10.1145/2907070},
	doi = {10.1145/2907070},
	number = {2},
	urldate = {2023-09-23},
	journal = {ACM Computing Surveys},
	author = {Branco, Paula and Torgo, Luís and Ribeiro, Rita P.},
	month = aug,
	year = {2016},
	pages = {31:1--31:50},
}

@misc{chen_ucr_2015,
	title = {The {UCR} {Time} {Series} {Classification} {Archive}},
	url = {www.cs.ucr.edu/~eamonn/time_series_data/},
	author = {Chen, Yanping and Keogh, Eamonn and Hu, Bing and Begum, Nurjahan and Bagnall, Anthony and Mueen, Abdullah and Batista, Gustavo},
	month = jul,
	year = {2015},
}

@article{sun_low-degree_2022,
	title = {Low-degree term first in {ResNet}, its variants and the whole neural network family},
	volume = {148},
	issn = {0893-6080},
	url = {https://www.sciencedirect.com/science/article/pii/S0893608022000120},
	doi = {10.1016/j.neunet.2022.01.012},
	urldate = {2023-09-20},
	journal = {Neural Networks},
	author = {Sun, Tongfeng and Ding, Shifei and Guo, Lili},
	month = apr,
	year = {2022},
	pages = {155--165},
}

@inproceedings{he_deep_2016,
	address = {Las Vegas, NV, USA},
	title = {Deep {Residual} {Learning} for {Image} {Recognition}},
	isbn = {978-1-4673-8851-1},
	url = {http://ieeexplore.ieee.org/document/7780459/},
	doi = {10.1109/CVPR.2016.90},
	language = {en},
	urldate = {2023-09-20},
	booktitle = {2016 {IEEE} {Conference} on {Computer} {Vision} and {Pattern} {Recognition} ({CVPR})},
	publisher = {IEEE},
	author = {He, Kaiming and Zhang, Xiangyu and Ren, Shaoqing and Sun, Jian},
	month = jun,
	year = {2016},
	pages = {770--778},
}

@article{cuppens_accelerometry-based_2014,
	title = {Accelerometry-based home monitoring for detection of nocturnal hypermotor seizures based on novelty detection},
	volume = {18},
	issn = {2168-2208},
	doi = {10.1109/JBHI.2013.2285015},
	language = {eng},
	number = {3},
	journal = {IEEE journal of biomedical and health informatics},
	author = {Cuppens, Kris and Karsmakers, Peter and Van de Vel, Anouk and Bonroy, Bert and Milosevic, Milica and Luca, Stijn and Croonenborghs, Tom and Ceulemans, Berten and Lagae, Lieven and Van Huffel, Sabine and Vanrumste, Bart},
	month = may,
	year = {2014},
	pmid = {24122607},
	pages = {1026--1033},
}

@article{meisel_machine_2020,
	title = {Machine learning from wristband sensor data for wearable, noninvasive seizure forecasting},
	volume = {61},
	copyright = {© 2020 The Authors. Epilepsia published by Wiley Periodicals LLC on behalf of International League Against Epilepsy},
	issn = {1528-1167},
	url = {https://onlinelibrary.wiley.com/doi/abs/10.1111/epi.16719},
	doi = {10.1111/epi.16719},
	language = {en},
	number = {12},
	urldate = {2023-09-19},
	journal = {Epilepsia},
	author = {Meisel, Christian and El Atrache, Rima and Jackson, Michele and Schubach, Sarah and Ufongene, Claire and Loddenkemper, Tobias},
	year = {2020},
	note = {\_eprint: https://onlinelibrary.wiley.com/doi/pdf/10.1111/epi.16719},
	pages = {2653--2666},
}

@article{yamakawaWear2020Anomaly,
	title = {Wearable {Epileptic} {Seizure} {Prediction} {System} with {Machine}-{Learning}-{Based} {Anomaly} {Detection} of {Heart} {Rate} {Variability}},
	volume = {20},
	copyright = {http://creativecommons.org/licenses/by/3.0/},
	issn = {1424-8220},
	url = {https://www.mdpi.com/1424-8220/20/14/3987},
	doi = {10.3390/s20143987},
	language = {en},
	number = {14},
	urldate = {2023-09-19},
	journal = {Sensors},
	author = {Yamakawa, Toshitaka and Miyajima, Miho and Fujiwara, Koichi and Kano, Manabu and Suzuki, Yoko and Watanabe, Yutaka and Watanabe, Satsuki and Hoshida, Tohru and Inaji, Motoki and Maehara, Taketoshi},
	month = jan,
	year = {2020},
	note = {Number: 14
Publisher: Multidisciplinary Digital Publishing Institute},
	pages = {3987},
}

@article{baghersalimi_personalized_2022,
	title = {Personalized {Real}-{Time} {Federated} {Learning} for {Epileptic} {Seizure} {Detection}},
	volume = {26},
	issn = {2168-2208},
	doi = {10.1109/JBHI.2021.3096127},
	number = {2},
	journal = {IEEE Journal of Biomedical and Health Informatics},
	author = {Baghersalimi, Saleh and Teijeiro, Tomas and Atienza, David and Aminifar, Amir},
	month = feb,
	year = {2022},
	note = {Conference Name: IEEE Journal of Biomedical and Health Informatics},
	pages = {898--909},
}

@article{you_semi-supervised_2022,
	title = {Semi-supervised automatic seizure detection using personalized anomaly detecting variational autoencoder with behind-the-ear {EEG}},
	volume = {213},
	issn = {0169-2607},
	url = {https://www.sciencedirect.com/science/article/pii/S0169260721006167},
	doi = {10.1016/j.cmpb.2021.106542},
	urldate = {2023-09-19},
	journal = {Computer Methods and Programs in Biomedicine},
	author = {You, Sungmin and Hwan Cho, Baek and Shon, Young-Min and Seo, Dae-Won and Kim, In Young},
	month = jan,
	year = {2022},
	pages = {106542},
}

@article{you_unsupervised_2020,
	title = {Unsupervised automatic seizure detection for focal-onset seizures recorded with behind-the-ear {EEG} using an anomaly-detecting generative adversarial network},
	volume = {193},
	issn = {0169-2607},
	url = {https://www.sciencedirect.com/science/article/pii/S0169260719320000},
	doi = {10.1016/j.cmpb.2020.105472},
	urldate = {2023-09-19},
	journal = {Computer Methods and Programs in Biomedicine},
	author = {You, Sungmin and Cho, Baek Hwan and Yook, Soonhyun and Kim, Joo Young and Shon, Young-Min and Seo, Dae-Won and Kim, In Young},
	month = sep,
	year = {2020},
	pages = {105472},
}

@article{bottcher_intra-_2022,
	title = {Intra- and {Inter}-{Subject} {Perspectives} on the {Detection} of {Focal} {Onset} {Motor} {Seizures} in {Epilepsy} {Patients}},
	volume = {22},
	issn = {1424-8220},
	url = {https://www.ncbi.nlm.nih.gov/pmc/articles/PMC9105312/},
	doi = {10.3390/s22093318},
	number = {9},
	urldate = {2023-09-18},
	journal = {Sensors (Basel, Switzerland)},
	author = {Böttcher, Sebastian and Bruno, Elisa and Epitashvili, Nino and Dümpelmann, Matthias and Zabler, Nicolas and Glasstetter, Martin and Ticcinelli, Valentina and Thorpe, Sarah and Lees, Simon and Van Laerhoven, Kristof and Richardson, Mark P. and Schulze-Bonhage, Andreas},
	month = apr,
	year = {2022},
	pmid = {35591007},
	pmcid = {PMC9105312},
	pages = {3318},
}

@article{bottcher_detecting_2021,
	title = {Detecting {Tonic}-{Clonic} {Seizures} in {Multimodal} {Biosignal} {Data} {From} {Wearables}: {Methodology} {Design} and {Validation}},
	volume = {9},
	issn = {2291-5222},
	shorttitle = {Detecting {Tonic}-{Clonic} {Seizures} in {Multimodal} {Biosignal} {Data} {From} {Wearables}},
	url = {https://www.ncbi.nlm.nih.gov/pmc/articles/PMC8663471/},
	doi = {10.2196/27674},
	number = {11},
	urldate = {2023-09-18},
	journal = {JMIR mHealth and uHealth},
	author = {Böttcher, Sebastian and Bruno, Elisa and Manyakov, Nikolay V and Epitashvili, Nino and Claes, Kasper and Glasstetter, Martin and Thorpe, Sarah and Lees, Simon and Dümpelmann, Matthias and Van Laerhoven, Kristof and Richardson, Mark P and Schulze-Bonhage, Andreas},
	month = nov,
	year = {2021},
	pmid = {34806993},
	pmcid = {PMC8663471},
	pages = {e27674},
}

@article{vandecasteele_power_2021,
	title = {The power of {ECG} in multimodal patient-specific seizure monitoring: {Added} value to an {EEG}-based detector using limited channels},
	volume = {62},
	issn = {1528-1167},
	shorttitle = {The power of {ECG} in multimodal patient-specific seizure monitoring},
	url = {https://onlinelibrary.wiley.com/doi/abs/10.1111/epi.16990},
	doi = {10.1111/epi.16990},
	language = {en},
	number = {10},
	urldate = {2023-09-18},
	journal = {Epilepsia},
	author = {Vandecasteele, Kaat and De Cooman, Thomas and Chatzichristos, Christos and Cleeren, Evy and Swinnen, Lauren and Macea Ortiz, Jaiver and Van Huffel, Sabine and Dümpelmann, Matthias and Schulze-Bonhage, Andreas and De Vos, Maarten and Van Paesschen, Wim and Hunyadi, Borbála},
	year = {2021},
	note = {\_eprint: https://onlinelibrary.wiley.com/doi/pdf/10.1111/epi.16990},
	pages = {2333--2343},
}

@article{forooghifar_self-aware_2022,
	title = {A {Self}-{Aware} {Epilepsy} {Monitoring} {System} for {Real}-{Time} {Epileptic} {Seizure} {Detection}},
	volume = {27},
	issn = {1383-469X, 1572-8153},
	url = {https://link.springer.com/10.1007/s11036-019-01322-7},
	doi = {10.1007/s11036-019-01322-7},
	language = {en},
	number = {2},
	urldate = {2023-09-18},
	journal = {Mobile Networks and Applications},
	author = {Forooghifar, Farnaz and Aminifar, Amir and Cammoun, Leila and Wisniewski, Ilona and Ciumas, Carolina and Ryvlin, Philippe and Atienza, David},
	month = apr,
	year = {2022},
	pages = {677--690},
}

@article{onorati_multicenter_2017,
	title = {Multicenter clinical assessment of improved wearable multimodal convulsive seizure detectors},
	volume = {58},
	copyright = {Wiley Periodicals, Inc. © 2017 International League Against Epilepsy},
	issn = {1528-1167},
	url = {https://onlinelibrary.wiley.com/doi/abs/10.1111/epi.13899},
	doi = {10.1111/epi.13899},
	language = {en},
	number = {11},
	urldate = {2023-09-18},
	journal = {Epilepsia},
	author = {Onorati, Francesco and Regalia, Giulia and Caborni, Chiara and Migliorini, Matteo and Bender, Daniel and Poh, Ming-Zher and Frazier, Cherise and Kovitch Thropp, Eliana and Mynatt, Elizabeth D. and Bidwell, Jonathan and Mai, Roberto and LaFrance Jr, W. Curt and Blum, Andrew S. and Friedman, Daniel and Loddenkemper, Tobias and Mohammadpour-Touserkani, Fatemeh and Reinsberger, Claus and Tognetti, Simone and Picard, Rosalind W.},
	year = {2017},
	note = {\_eprint: https://onlinelibrary.wiley.com/doi/pdf/10.1111/epi.13899},
	pages = {1870--1879},
}

@inproceedings{heldberg_using_2015,
	title = {Using wearable sensors for semiology-independent seizure detection - towards ambulatory monitoring of epilepsy},
	doi = {10.1109/EMBC.2015.7319660},
	booktitle = {2015 37th {Annual} {International} {Conference} of the {IEEE} {Engineering} in {Medicine} and {Biology} {Society} ({EMBC})},
	author = {Heldberg, Beeke E. and Kautz, Thomas and Leutheuser, Heike and Hopfengärtner, Rüdiger and Kasper, Burkhard S. and Eskofier, Bjoern M.},
	month = aug,
	year = {2015},
	note = {ISSN: 1558-4615},
	pages = {5593--5596},
}

@article{bruno_wearable_2018,
	title = {Wearable technology in epilepsy: {The} views of patients, caregivers, and healthcare professionals},
	volume = {85},
	issn = {1525-5050},
	shorttitle = {Wearable technology in epilepsy},
	url = {https://www.sciencedirect.com/science/article/pii/S1525505018303561},
	doi = {10.1016/j.yebeh.2018.05.044},
	urldate = {2023-09-14},
	journal = {Epilepsy \& Behavior},
	author = {Bruno, Elisa and Simblett, Sara and Lang, Alexandra and Biondi, Andrea and Odoi, Clarissa and Schulze-Bonhage, Andreas and Wykes, Til and Richardson, Mark P.},
	month = aug,
	year = {2018},
	pages = {141--149},
}

@article{poh_autonomic_2012,
	title = {Autonomic changes with seizures correlate with postictal {EEG} suppression},
	volume = {78},
	copyright = {Copyright © 2012 by AAN Enterprises, Inc.},
	issn = {0028-3878, 1526-632X},
	url = {https://n.neurology.org/content/78/23/1868},
	doi = {10.1212/WNL.0b013e318258f7f1},
	language = {en},
	number = {23},
	urldate = {2023-09-11},
	journal = {Neurology},
	author = {Poh, M.-Z. and Loddenkemper, T. and Reinsberger, C. and Swenson, N. C. and Goyal, S. and Madsen, J. R. and Picard, R. W.},
	month = jun,
	year = {2012},
	pmid = {22539579},
	note = {Publisher: Wolters Kluwer Health, Inc. on behalf of the American Academy of Neurology
Section: Articles},
	pages = {1868--1876},
}

@inproceedings{poh_continuous_2010,
	title = {Continuous monitoring of electrodermal activity during epileptic seizures using a wearable sensor},
	doi = {10.1109/IEMBS.2010.5625988},
	booktitle = {2010 {Annual} {International} {Conference} of the {IEEE} {Engineering} in {Medicine} and {Biology}},
	author = {Poh, Ming-Zher and Loddenkemper, Tobias and Swenson, Nicholas C. and Goyal, Shubhi and Madsen, Joseph R. and Picard, Rosalind W.},
	month = aug,
	year = {2010},
	note = {ISSN: 1558-4615},
	pages = {4415--4418},
}

@article{beniczky_non-electroencephalography-based_2019,
	title = {Non-electroencephalography-based seizure detection},
	volume = {32},
	issn = {1350-7540},
	url = {https://journals.lww.com/co-neurology/abstract/2019/04000/non_electroencephalography_based_seizure_detection.5.aspx},
	doi = {10.1097/WCO.0000000000000658},
	language = {en-US},
	number = {2},
	urldate = {2023-09-11},
	journal = {Current Opinion in Neurology},
	author = {Beniczky, Sándor and Jeppesen, Jesper},
	month = apr,
	year = {2019},
	pages = {198},
}

@article{ulate-campos_automated_2016,
	title = {Automated seizure detection systems and their effectiveness for each type of seizure},
	volume = {40},
	issn = {1059-1311},
	url = {https://www.sciencedirect.com/science/article/pii/S1059131116300711},
	doi = {10.1016/j.seizure.2016.06.008},
	urldate = {2023-09-11},
	journal = {Seizure},
	author = {Ulate-Campos, A. and Coughlin, F. and Gaínza-Lein, M. and Fernández, I. Sánchez and Pearl, P. L. and Loddenkemper, T.},
	month = aug,
	year = {2016},
	pages = {88--101},
}

@article{baumgartner_using_2021,
	title = {Using {sEMG} to identify seizure semiology of motor seizures},
	volume = {86},
	issn = {1059-1311},
	url = {https://www.sciencedirect.com/science/article/pii/S1059131120303654},
	doi = {10.1016/j.seizure.2020.11.013},
	urldate = {2023-09-11},
	journal = {Seizure},
	author = {Baumgartner, Christoph and Whitmire, Luke E. and Voyles, Shannon R. and Cardenas, Damon P.},
	month = mar,
	year = {2021},
	pages = {52--59},
}

@article{arends_multimodal_2018,
	title = {Multimodal nocturnal seizure detection in a residential care setting: {A} long-term prospective trial},
	volume = {91},
	copyright = {Copyright © 2018 The Author(s). Published by Wolters Kluwer Health, Inc. on behalf of the American Academy of Neurology.. This is an open access article distributed under the terms of the Creative Commons Attribution-NonCommercial-NoDerivatives License 4.0 (CC BY-NC-ND), which permits downloading and sharing the work provided it is properly cited. The work cannot be changed in any way or used commercially without permission from the journal.},
	issn = {0028-3878, 1526-632X},
	shorttitle = {Multimodal nocturnal seizure detection in a residential care setting},
	url = {https://n.neurology.org/content/91/21/e2010},
	doi = {10.1212/WNL.0000000000006545},
	language = {en},
	number = {21},
	urldate = {2023-09-11},
	journal = {Neurology},
	author = {Arends, Johan and Thijs, Roland D. and Gutter, Thea and Ungureanu, Constantin and Cluitmans, Pierre and Dijk, Johannes Van and Andel, Judith van and Tan, Francis and Weerd, Al de and Vledder, Ben and Hofstra, Wytske and Lazeron, Richard and Thiel, Ghislaine van and Roes, Kit C. B. and Leijten, Frans and Consortium, {and} the Dutch Tele-Epilepsy},
	month = nov,
	year = {2018},
	pmid = {30355702},
	note = {Publisher: Wolters Kluwer Health, Inc. on behalf of the American Academy of Neurology
Section: Article},
	pages = {e2010--e2019},
}

@article{glasstetter_identification_2021,
	title = {Identification of {Ictal} {Tachycardia} in {Focal} {Motor}- and {Non}-{Motor} {Seizures} by {Means} of a {Wearable} {PPG} {Sensor}},
	volume = {21},
	issn = {1424-8220},
	url = {https://www.ncbi.nlm.nih.gov/pmc/articles/PMC8470979/},
	doi = {10.3390/s21186017},
	number = {18},
	urldate = {2023-09-11},
	journal = {Sensors (Basel, Switzerland)},
	author = {Glasstetter, Martin and Böttcher, Sebastian and Zabler, Nicolas and Epitashvili, Nino and Dümpelmann, Matthias and Richardson, Mark P. and Schulze-Bonhage, Andreas},
	month = sep,
	year = {2021},
	pmid = {34577222},
	pmcid = {PMC8470979},
	pages = {6017},
}

@article{el_atrache_photoplethysmography_2020,
	title = {Photoplethysmography: {A} measure for the function of the autonomic nervous system in focal impaired awareness seizures},
	volume = {61},
	issn = {1528-1167},
	shorttitle = {Photoplethysmography},
	url = {https://onlinelibrary.wiley.com/doi/abs/10.1111/epi.16621},
	doi = {10.1111/epi.16621},
	language = {en},
	number = {8},
	urldate = {2023-09-11},
	journal = {Epilepsia},
	author = {El Atrache, Rima and Tamilia, Eleonora and Mohammadpour Touserkani, Fatemeh and Hammond, Sarah and Papadelis, Christos and Kapur, Kush and Jackson, Michele and Bucciarelli, Bethany and Tsuboyama, Melissa and Sarkis, Rani A. and Loddenkemper, Tobias},
	year = {2020},
	note = {\_eprint: https://onlinelibrary.wiley.com/doi/pdf/10.1111/epi.16621},
	pages = {1617--1626},
}

@article{vandecasteele_automated_2017,
	title = {Automated {Epileptic} {Seizure} {Detection} {Based} on {Wearable} {ECG} and {PPG} in a {Hospital} {Environment}},
	volume = {17},
	copyright = {http://creativecommons.org/licenses/by/3.0/},
	issn = {1424-8220},
	url = {https://www.mdpi.com/1424-8220/17/10/2338},
	doi = {10.3390/s17102338},
	language = {en},
	number = {10},
	urldate = {2023-09-11},
	journal = {Sensors},
	author = {Vandecasteele, Kaat and De Cooman, Thomas and Gu, Ying and Cleeren, Evy and Claes, Kasper and Paesschen, Wim Van and Huffel, Sabine Van and Hunyadi, Borbála},
	month = oct,
	year = {2017},
	note = {Number: 10
Publisher: Multidisciplinary Digital Publishing Institute},
	pages = {2338},
}

@article{van_andel_multimodal_2017,
	title = {Multimodal, automated detection of nocturnal motor seizures at home: {Is} a reliable seizure detector feasible?},
	volume = {2},
	issn = {2470-9239},
	shorttitle = {Multimodal, automated detection of nocturnal motor seizures at home},
	doi = {10.1002/epi4.12076},
	language = {eng},
	number = {4},
	journal = {Epilepsia Open},
	author = {van Andel, Judith and Ungureanu, Constantin and Arends, Johan and Tan, Francis and Van Dijk, Johannes and Petkov, George and Kalitzin, Stiliyan and Gutter, Thea and de Weerd, Al and Vledder, Ben and Thijs, Roland and van Thiel, Ghislaine and Roes, Kit and Leijten, Frans},
	month = dec,
	year = {2017},
	pmid = {29588973},
	pmcid = {PMC5862103},
	pages = {424--431},
}

@article{jeppesen_detection_2015,
	title = {Detection of epileptic seizures with a modified heart rate variability algorithm based on {Lorenz} plot},
	volume = {24},
	issn = {1059-1311},
	url = {https://www.sciencedirect.com/science/article/pii/S1059131114003033},
	doi = {10.1016/j.seizure.2014.11.004},
	urldate = {2023-09-11},
	journal = {Seizure},
	author = {Jeppesen, Jesper and Beniczky, Sándor and Johansen, Peter and Sidenius, Per and Fuglsang-Frederiksen, Anders},
	month = jan,
	year = {2015},
	pages = {1--7},
}

@article{van_elmpt_model_2006,
	title = {A model of heart rate changes to detect seizures in severe epilepsy},
	volume = {15},
	issn = {1059-1311},
	doi = {10.1016/j.seizure.2006.03.005},
	language = {eng},
	number = {6},
	journal = {Seizure},
	author = {van Elmpt, Wouter J. C. and Nijsen, Tamara M. E. and Griep, Paul A. M. and Arends, Johan B. A. M.},
	month = sep,
	year = {2006},
	pmid = {16828317},
	pages = {366--375},
}

@article{jahanbekam_performance_2021,
	title = {Performance of {ECG}-based seizure detection algorithms strongly depends on training and test conditions},
	volume = {6},
	issn = {2470-9239},
	doi = {10.1002/epi4.12520},
	language = {eng},
	number = {3},
	journal = {Epilepsia Open},
	author = {Jahanbekam, Amirhossein and Baumann, Jan and Nass, Robert D. and Bauckhage, Christian and Hill, Holger and Elger, Christian E. and Surges, Rainer},
	month = sep,
	year = {2021},
	pmid = {34250754},
	pmcid = {PMC8408591},
	pages = {597--606},
}

@article{fanghyChenSeizuresDetectionReview2022,
	title = {Seizures detection using multimodal signals: a scoping review},
	volume = {43},
	issn = {1361-6579},
	shorttitle = {Seizures detection using multimodal signals},
	doi = {10.1088/1361-6579/ac7a8d},
	language = {eng},
	number = {7},
	journal = {Physiological Measurement},
	author = {Chen, Fangyi and Chen, Ina and Zafar, Muhammad and Sinha, Saurabh R. and Hu, Xiao},
	month = jul,
	year = {2022},
	pmid = {35724654},
}

@article{brinkmannSeizureDiariesReview2021,
	title = {Seizure {Diaries} and {Forecasting} {With} {Wearables}: {Epilepsy} {Monitoring} {Outside} the {Clinic}},
	volume = {12},
	issn = {1664-2295},
	shorttitle = {Seizure {Diaries} and {Forecasting} {With} {Wearables}},
	doi = {10.3389/fneur.2021.690404},
	language = {eng},
	journal = {Frontiers in Neurology},
	author = {Brinkmann, Benjamin H. and Karoly, Philippa J. and Nurse, Ewan S. and Dumanis, Sonya B. and Nasseri, Mona and Viana, Pedro F. and Schulze-Bonhage, Andreas and Freestone, Dean R. and Worrell, Greg and Richardson, Mark P. and Cook, Mark J.},
	year = {2021},
	pmid = {34326807},
	pmcid = {PMC8315760},
	pages = {690404},
}

@inproceedings{gheryani_epileptic_2020,
	title = {Epileptic {Seizures} {Detection} based on {Inertial} and {Physiological} {Data} from {Wireless} {Body} {Sensors}},
	doi = {10.1109/ICC40277.2020.9148676},
	booktitle = {{ICC} 2020 - 2020 {IEEE} {International} {Conference} on {Communications} ({ICC})},
	author = {Gheryani, Mostafa and Salem, Osman and Mehaoua, Ahmed},
	month = jun,
	year = {2020},
	note = {ISSN: 1938-1883},
	pages = {1--6},
}

@article{nasseri_non-invasive_2021,
	title = {Non-invasive wearable seizure detection using long-short-term memory networks with transfer learning},
	volume = {18},
	issn = {1741-2552},
	doi = {10.1088/1741-2552/abef8a},
	language = {eng},
	number = {5},
	journal = {Journal of Neural Engineering},
	author = {Nasseri, Mona and Pal Attia, Tal and Joseph, Boney and Gregg, Nicholas M. and Nurse, Ewan S. and Viana, Pedro F. and Schulze-Bonhage, Andreas and Dümpelmann, Matthias and Worrell, Gregory and Freestone, Dean R. and Richardson, Mark P. and Brinkmann, Benjamin H.},
	month = apr,
	year = {2021},
	pmid = {33730713},
}

@article{munch_nielsen_towards_2022,
	title = {Towards a wearable multi-modal seizure detection system in epilepsy: {A} pilot study},
	volume = {136},
	issn = {1872-8952},
	shorttitle = {Towards a wearable multi-modal seizure detection system in epilepsy},
	doi = {10.1016/j.clinph.2022.01.005},
	language = {eng},
	journal = {Clinical Neurophysiology: Official Journal of the International Federation of Clinical Neurophysiology},
	author = {Munch Nielsen, Jonas and Zibrandtsen, Ivan C. and Masulli, Paolo and Lykke Sørensen, Torben and Andersen, Tobias S. and Wesenberg Kjær, Troels},
	month = apr,
	year = {2022},
	pmid = {35131637},
	pages = {40--48},
}

@article{hegarty-craver_cardiac-based_2021,
	title = {Cardiac-based detection of seizures in children with epilepsy},
	volume = {122},
	issn = {1525-5069},
	doi = {10.1016/j.yebeh.2021.108129},
	language = {eng},
	journal = {Epilepsy \& Behavior: E\&B},
	author = {Hegarty-Craver, Meghan and Kroner, Barbara L. and Bumbut, Adrian and DeFilipp, Samuel J. and Gaillard, William D. and Gilchrist, Kristin H.},
	month = sep,
	year = {2021},
	pmid = {34147021},
	pmcid = {PMC8429110},
	pages = {108129},
}

@article{giannakakis_focal_2019,
	title = {Focal epileptic seizures anticipation based on patterns of heart rate variability parameters},
	volume = {178},
	issn = {1872-7565},
	doi = {10.1016/j.cmpb.2019.05.032},
	language = {eng},
	journal = {Computer Methods and Programs in Biomedicine},
	author = {Giannakakis, Giorgos and Tsiknakis, Manolis and Vorgia, Pelagia},
	month = sep,
	year = {2019},
	pmid = {31416541},
	pages = {123--133},
}

@article{jeppesen_seizure_2020,
	title = {Seizure detection using heart rate variability: {A} prospective validation study},
	volume = {61},
	copyright = {© 2020 International League Against Epilepsy},
	issn = {1528-1167},
	shorttitle = {Seizure detection using heart rate variability},
	url = {https://onlinelibrary.wiley.com/doi/abs/10.1111/epi.16511},
	doi = {10.1111/epi.16511},
	language = {en},
	number = {S1},
	urldate = {2023-09-09},
	journal = {Epilepsia},
	author = {Jeppesen, Jesper and Fuglsang-Frederiksen, Anders and Johansen, Peter and Christensen, Jakob and Wüstenhagen, Stephan and Tankisi, Hatice and Qerama, Erisela and Beniczky, Sándor},
	year = {2020},
	note = {\_eprint: https://onlinelibrary.wiley.com/doi/pdf/10.1111/epi.16511},
	pages = {S41--S46},
}

@article{jeppesen_seizure_2019,
	title = {Seizure detection based on heart rate variability using a wearable electrocardiography device},
	volume = {60},
	issn = {1528-1167},
	doi = {10.1111/epi.16343},
	language = {eng},
	number = {10},
	journal = {Epilepsia},
	author = {Jeppesen, Jesper and Fuglsang-Frederiksen, Anders and Johansen, Peter and Christensen, Jakob and Wüstenhagen, Stephan and Tankisi, Hatice and Qerama, Erisela and Hess, Alexander and Beniczky, Sándor},
	month = oct,
	year = {2019},
	pmid = {31538347},
	pages = {2105--2113},
}

@article{johansson_tonic-clonic_2019,
	title = {Tonic-clonic seizure detection using accelerometry-based wearable sensors: {A} prospective, video-{EEG} controlled study},
	volume = {65},
	issn = {1532-2688},
	shorttitle = {Tonic-clonic seizure detection using accelerometry-based wearable sensors},
	doi = {10.1016/j.seizure.2018.12.024},
	language = {eng},
	journal = {Seizure},
	author = {Johansson, Dongni and Ohlsson, Fredrik and Krýsl, David and Rydenhag, Bertil and Czarnecki, Madeleine and Gustafsson, Niclas and Wipenmyr, Jan and McKelvey, Tomas and Malmgren, Kristina},
	month = feb,
	year = {2019},
	pmid = {30611010},
	pages = {48--54},
}

@article{fisher_bad_2017,
	title = {Bad information in epilepsy care},
	volume = {67},
	issn = {1525-5050},
	url = {https://www.sciencedirect.com/science/article/pii/S1525505016305765},
	doi = {10.1016/j.yebeh.2016.10.022},
	urldate = {2023-09-09},
	journal = {Epilepsy \& Behavior},
	author = {Fisher, Robert S.},
	month = feb,
	year = {2017},
	pages = {133--134},
}

@article{lanzone_transient_2018,
	title = {Transient epileptic and global amnesia: {Real}-life differential diagnosis},
	volume = {88},
	issn = {1525-5050},
	shorttitle = {Transient epileptic and global amnesia},
	url = {https://www.sciencedirect.com/science/article/pii/S1525505018302130},
	doi = {10.1016/j.yebeh.2018.07.015},
	urldate = {2023-09-09},
	journal = {Epilepsy \& Behavior},
	author = {Lanzone, Jacopo and Ricci, Lorenzo and Assenza, Giovanni and Ulivi, Martina and Di Lazzaro, Vincenzo and Tombini, Mario},
	month = nov,
	year = {2018},
	pages = {205--211},
}

@article{espinosa-jovel_epidemiological_2018,
	title = {Epidemiological profile of epilepsy in low income populations},
	volume = {56},
	issn = {1059-1311},
	url = {https://www.sciencedirect.com/science/article/pii/S1059131117306568},
	doi = {10.1016/j.seizure.2018.02.002},
	urldate = {2023-09-08},
	journal = {Seizure},
	author = {Espinosa-Jovel, Camilo and Toledano, Rafael and Aledo-Serrano, Ángel and García-Morales, Irene and Gil-Nagel, Antonio},
	month = mar,
	year = {2018},
	pages = {67--72},
}

@article{shorvon_sudden_2011,
	title = {Sudden unexpected death in epilepsy},
	volume = {378},
	issn = {0140-6736},
	url = {https://www.sciencedirect.com/science/article/pii/S0140673611601761},
	doi = {10.1016/S0140-6736(11)60176-1},
	number = {9808},
	urldate = {2023-09-08},
	journal = {The Lancet},
	author = {Shorvon, Simon and Tomson, Torbjorn},
	month = dec,
	year = {2011},
	pages = {2028--2038},
}

@article{trinkaStatusEpileticus2015,
	title = {A definition and classification of status epilepticus – {Report} of the {ILAE} {Task} {Force} on {Classification} of {Status} {Epilepticus}},
	volume = {56},
	copyright = {Wiley Periodicals, Inc. © 2015 International League Against Epilepsy},
	issn = {1528-1167},
	url = {https://onlinelibrary.wiley.com/doi/abs/10.1111/epi.13121},
	doi = {10.1111/epi.13121},
	language = {en},
	number = {10},
	urldate = {2023-09-06},
	journal = {Epilepsia},
	author = {Trinka, Eugen and Cock, Hannah and Hesdorffer, Dale and Rossetti, Andrea O. and Scheffer, Ingrid E. and Shinnar, Shlomo and Shorvon, Simon and Lowenstein, Daniel H.},
	year = {2015},
	note = {\_eprint: https://onlinelibrary.wiley.com/doi/pdf/10.1111/epi.13121},
	pages = {1515--1523},
}

@article{wenLiReview2022,
	title = {Seizure detection based on wearable devices: {A} review of device, mechanism, and algorithm},
	volume = {146},
	copyright = {© 2022 John Wiley \& Sons A/S. Published by John Wiley \& Sons Ltd.},
	issn = {1600-0404},
	shorttitle = {Seizure detection based on wearable devices},
	url = {https://onlinelibrary.wiley.com/doi/abs/10.1111/ane.13716},
	doi = {10.1111/ane.13716},
	language = {en},
	number = {6},
	urldate = {2023-09-06},
	journal = {Acta Neurologica Scandinavica},
	author = {Li, Wen and Wang, Guangming and Lei, Xiyuan and Sheng, Duozheng and Yu, Tao and Wang, Gang},
	year = {2022},
	note = {\_eprint: https://onlinelibrary.wiley.com/doi/pdf/10.1111/ane.13716},
	pages = {723--731},
}

@article{50millionGlobally,
	title = {Status of epilepsy in the tropics: {An} overlooked perspective},
	volume = {8},
	issn = {2470-9239},
	shorttitle = {Status of epilepsy in the tropics},
	url = {https://www.ncbi.nlm.nih.gov/pmc/articles/PMC9977758/},
	doi = {10.1002/epi4.12686},
	number = {1},
	urldate = {2023-09-06},
	journal = {Epilepsia Open},
	author = {Liu, Jiaqi and Zhang, Peng and Zou, Qin and Liang, Jiantang and Chen, Yongmin and Cai, Yi and Li, Shichuo and Li, Jinmei and Su, Jing and Li, Qifu},
	month = jan,
	year = {2023},
	pmid = {36588194},
	pmcid = {PMC9977758},
	pages = {32--45},
}

@article{poh2012,
	title = {Convulsive seizure detection using a wrist-worn electrodermal activity and accelerometry biosensor},
	volume = {53},
	copyright = {Wiley Periodicals, Inc. © 2012 International League Against Epilepsy},
	issn = {1528-1167},
	url = {https://onlinelibrary.wiley.com/doi/abs/10.1111/j.1528-1167.2012.03444.x},
	doi = {10.1111/j.1528-1167.2012.03444.x},
	language = {en},
	number = {5},
	urldate = {2023-09-05},
	journal = {Epilepsia},
	author = {Poh, Ming-Zher and Loddenkemper, Tobias and Reinsberger, Claus and Swenson, Nicholas C. and Goyal, Shubhi and Sabtala, Mangwe C. and Madsen, Joseph R. and Picard, Rosalind W.},
	year = {2012},
	note = {\_eprint: https://onlinelibrary.wiley.com/doi/pdf/10.1111/j.1528-1167.2012.03444.x},
	pages = {e93--e97},
}

@article{milosevic2016,
	title = {Automated {Detection} of {Tonic}–{Clonic} {Seizures} {Using} 3-{D} {Accelerometry} and {Surface} {Electromyography} in {Pediatric} {Patients}},
	volume = {20},
	issn = {2168-2208},
	doi = {10.1109/JBHI.2015.2462079},
	number = {5},
	journal = {IEEE Journal of Biomedical and Health Informatics},
	author = {Milošević, Milica and Van de Vel, Anouk and Bonroy, Bert and Ceulemans, Berten and Lagae, Lieven and Vanrumste, Bart and Huffel, Sabine Van},
	month = sep,
	year = {2016},
	note = {Conference Name: IEEE Journal of Biomedical and Health Informatics},
	pages = {1333--1341},
}

@article{agrahriAnushman2022,
	title = {Detection of {Epileptic} {Seizure} {Using} {Accelerometer} {Time} {Series} {Data} and {Hidden} {Markov} {Model}},
	volume = {2022},
	issn = {2694-0604},
	doi = {10.1109/EMBC48229.2022.9871914},
	language = {eng},
	journal = {Annual International Conference of the IEEE Engineering in Medicine and Biology Society. IEEE Engineering in Medicine and Biology Society. Annual International Conference},
	author = {Agrahri, Anshuman and Tyagi, Alok and Kumar, Dheeraj and Kusumakar, Shitanshu and Palaniswami, Marimuthu and Yan, Bernard},
	month = jul,
	year = {2022},
	pmid = {36086544},
	pages = {2426--2429},
}

@article{tangJianbin2021,
	title = {Seizure detection using wearable sensors and machine learning: {Setting} a benchmark},
	volume = {62},
	copyright = {© 2021 The Authors. Epilepsia published by Wiley Periodicals LLC on behalf of International League Against Epilepsy},
	issn = {1528-1167},
	shorttitle = {Seizure detection using wearable sensors and machine learning},
	url = {https://onlinelibrary.wiley.com/doi/abs/10.1111/epi.16967},
	doi = {10.1111/epi.16967},
	language = {en},
	number = {8},
	urldate = {2023-09-04},
	journal = {Epilepsia},
	author = {Tang, Jianbin and El Atrache, Rima and Yu, Shuang and Asif, Umar and Jackson, Michele and Roy, Subhrajit and Mirmomeni, Mahtab and Cantley, Sarah and Sheehan, Theodore and Schubach, Sarah and Ufongene, Claire and Vieluf, Solveig and Meisel, Christian and Harrer, Stefan and Loddenkemper, Tobias},
	year = {2021},
	note = {\_eprint: https://onlinelibrary.wiley.com/doi/pdf/10.1111/epi.16967},
	pages = {1807--1819},
}

@inproceedings{motaharErina2021,
	title = {Accelerometer-based {Convulsive} {Seizure} {Detection} using {CNN}},
	doi = {10.1109/BECITHCON54710.2021.9893602},
	booktitle = {2021 {IEEE} {International} {Conference} on {Biomedical} {Engineering}, {Computer} and {Information} {Technology} for {Health} ({BECITHCON})},
	author = {Motahar, Erina Binte and Ishtiaque, Farhan and Wadud, Md. Sharjis Ibne},
	month = dec,
	year = {2021},
	pages = {46--51},
}

@inproceedings{dongChunjiao2021,
	title = {Home-based {Detection} of {Epileptic} {Seizures} {Using} a {Bracelet} with {Motor} {Sensors}},
	doi = {10.1109/NER49283.2021.9441171},
	booktitle = {2021 10th {International} {IEEE}/{EMBS} {Conference} on {Neural} {Engineering} ({NER})},
	author = {Dong, Chunjiao and Chen, Lei and Ye, Tianchun and Long, Xi and Aarts, Ronald M. and van Dijk, Johannes and Shang, Chunheng and Liao, Xiwen and Wang, Yunfeng},
	month = may,
	year = {2021},
	note = {ISSN: 1948-3554},
	pages = {854--857},
}

@article{onorati,
	title = {Prospective {Study} of a {Multimodal} {Convulsive} {Seizure} {Detection} {Wearable} {System} on {Pediatric} and {Adult} {Patients} in the {Epilepsy} {Monitoring} {Unit}},
	volume = {12},
	issn = {1664-2295},
	url = {https://www.frontiersin.org/articles/10.3389/fneur.2021.724904},
	urldate = {2023-09-04},
	journal = {Frontiers in Neurology},
	author = {Onorati, Francesco and Regalia, Giulia and Caborni, Chiara and LaFrance, W. Curt and Blum, Andrew S. and Bidwell, Jonathan and De Liso, Paola and El Atrache, Rima and Loddenkemper, Tobias and Mohammadpour-Touserkani, Fatemeh and Sarkis, Rani A. and Friedman, Daniel and Jeschke, Jay and Picard, Rosalind},
	year = {2021},
}

@article{beniczky2018,
	title = {Automated real-time detection of tonic-clonic seizures using a wearable {EMG} device},
	volume = {90},
	copyright = {© 2018 The Author(s). Published by Wolters Kluwer Health, Inc. on behalf of the American Academy of Neurology.. This is an open access article distributed under the terms of the Creative Commons Attribution-NonCommercial-NoDerivatives License 4.0 (CC BY-NC-ND), which permits downloading and sharing the work provided it is properly cited. The work cannot be changed in any way or used commercially without permission from the journal.},
	issn = {0028-3878, 1526-632X},
	url = {https://n-neurology-org.eu1.proxy.openathens.net/content/90/5/e428},
	doi = {10.1212/WNL.0000000000004893},
	language = {en},
	number = {5},
	urldate = {2023-09-04},
	journal = {Neurology},
	author = {Beniczky, Sándor and Conradsen, Isa and Henning, Oliver and Fabricius, Martin and Wolf, Peter},
	month = jan,
	year = {2018},
	pmid = {29305441},
	note = {Publisher: Wolters Kluwer Health, Inc. on behalf of the American Academy of Neurology
Section: Article},
	pages = {e428--e434},
}

@article{szabo,
	title = {Electromyography-based seizure detector: {Preliminary} results comparing a generalized tonic–clonic seizure detection algorithm to video-{EEG} recordings},
	volume = {56},
	copyright = {© 2015 The Authors. Epilepsia published by Wiley Periodicals Inc. on behalf of International League Against Epilepsy.},
	issn = {1528-1167},
	shorttitle = {Electromyography-based seizure detector},
	url = {https://onlinelibrary.wiley.com/doi/abs/10.1111/epi.13083},
	doi = {10.1111/epi.13083},
	language = {en},
	number = {9},
	urldate = {2023-09-04},
	journal = {Epilepsia},
	author = {Szabó, Charles Ákos and Morgan, Lola C. and Karkar, Kameel M. and Leary, Linda D. and Lie, Octavian V. and Girouard, Michael and Cavazos, José E.},
	year = {2015},
	note = {\_eprint: https://onlinelibrary.wiley.com/doi/pdf/10.1111/epi.13083},
	pages = {1432--1437},
}

@article{beniczky,
	title = {Detection of generalized tonic–clonic seizures by a wireless wrist accelerometer: {A} prospective, multicenter study},
	volume = {54},
	copyright = {Wiley Periodicals, Inc. © 2013 International League Against Epilepsy},
	issn = {1528-1167},
	shorttitle = {Detection of generalized tonic–clonic seizures by a wireless wrist accelerometer},
	url = {https://onlinelibrary.wiley.com/doi/abs/10.1111/epi.12120},
	doi = {10.1111/epi.12120},
	language = {en},
	number = {4},
	urldate = {2023-09-04},
	journal = {Epilepsia},
	author = {Beniczky, Sándor and Polster, Tilman and Kjaer, Troels W. and Hjalgrim, Helle},
	year = {2013},
	note = {\_eprint: https://onlinelibrary.wiley.com/doi/pdf/10.1111/epi.12120},
	pages = {e58--e61},
}

@article{lockman,
	title = {Detection of seizure-like movements using a wrist accelerometer},
	volume = {20},
	issn = {1525-5050},
	url = {https://www.sciencedirect.com/science/article/pii/S1525505011000229},
	doi = {10.1016/j.yebeh.2011.01.019},
	number = {4},
	urldate = {2023-09-04},
	journal = {Epilepsy \& Behavior},
	author = {Lockman, Juliana and Fisher, Robert S. and Olson, Donald M.},
	month = apr,
	year = {2011},
	pages = {638--641},
}

@article{kramer,
	title = {A {Novel} {Portable} {Seizure} {Detection} {Alarm} {System}: {Preliminary} {Results}},
	volume = {28},
	issn = {0736-0258},
	shorttitle = {A {Novel} {Portable} {Seizure} {Detection} {Alarm} {System}},
	url = {https://journals.lww.com/clinicalneurophys/abstract/2011/02000/a_novel_portable_seizure_detection_alarm_system_.6.aspx},
	doi = {10.1097/WNP.0b013e3182051320},
	language = {en-US},
	number = {1},
	urldate = {2023-09-04},
	journal = {Journal of Clinical Neurophysiology},
	author = {Kramer, Uri and Kipervasser, Svetlana and Shlitner, Arie and Kuzniecky, Ruben},
	month = feb,
	year = {2011},
	pages = {36},
}

@article{conradsen,
	title = {Automated {Algorithm} for {Generalized} {Tonic}–{Clonic} {Epileptic} {Seizure} {Onset} {Detection} {Based} on {sEMG} {Zero}-{Crossing} {Rate}},
	volume = {59},
	issn = {1558-2531},
	doi = {10.1109/TBME.2011.2178094},
	number = {2},
	journal = {IEEE Transactions on Biomedical Engineering},
	author = {Conradsen, Isa and Beniczky, Sándor and Hoppe, Karsten and Wolf, Peter and Sorensen, Helge B. D.},
	month = feb,
	year = {2012},
	note = {Conference Name: IEEE Transactions on Biomedical Engineering},
	pages = {579--585},
}

@misc{resnet_TSC,
	title = {Time {Series} {Classification} from {Scratch} with {Deep} {Neural} {Networks}: {A} {Strong} {Baseline}},
	shorttitle = {Time {Series} {Classification} from {Scratch} with {Deep} {Neural} {Networks}},
	url = {http://arxiv.org/abs/1611.06455},
	urldate = {2023-09-04},
	publisher = {arXiv},
	author = {Wang, Zhiguang and Yan, Weizhong and Oates, Tim},
	month = dec,
	year = {2016},
	note = {arXiv:1611.06455 [cs, stat]},
}

@misc{he_deep_2015,
	title = {Deep {Residual} {Learning} for {Image} {Recognition}},
	url = {http://arxiv.org/abs/1512.03385},
	urldate = {2023-09-04},
	publisher = {arXiv},
	author = {He, Kaiming and Zhang, Xiangyu and Ren, Shaoqing and Sun, Jian},
	month = dec,
	year = {2015},
	note = {arXiv:1512.03385 [cs]},
}

@article{baseline,
	title = {Automated {Detection} of {Convulsive} {Seizures} {Using} a {Wearable} {Accelerometer} {Device}},
	volume = {66},
	issn = {1558-2531},
	doi = {10.1109/TBME.2018.2845865},
	number = {2},
	journal = {IEEE Transactions on Biomedical Engineering},
	author = {Kusmakar, Shitanshu and Karmakar, Chandan K. and Yan, Bernard and J.O’Brien, Terence and Muthuganapathy, Ramanathan and Palaniswami, Marimuthu},
	month = feb,
	year = {2019},
	note = {Conference Name: IEEE Transactions on Biomedical Engineering},
	pages = {421--432},
}

@article{commonSeizuresSurvay,
	title = {People with epilepsy are diagnosed most often with unspecified epilepsy, followed by focal epilepsy, generalized convulsive epilepsy, and generalized nonconvulsive epilepsy—{US} {MarketScan} data, 2010–2015},
	volume = {79},
	issn = {1525-5050},
	url = {https://www.sciencedirect.com/science/article/pii/S1525505017307187},
	doi = {10.1016/j.yebeh.2017.11.004},
	urldate = {2023-09-04},
	journal = {Epilepsy \& Behavior},
	author = {Sapkota, Sanjeeb and Kobau, Rosemarie and Pastula, Daniel M. and Zack, Matthew M.},
	month = feb,
	year = {2018},
	pages = {244--246},
}

@article{ahmadi_eeg-based_2020,
	title = {{EEG}-based classification of epilepsy and {PNES}: {EEG} microstate and functional brain network features},
	volume = {7},
	issn = {2198-4018},
	shorttitle = {{EEG}-based classification of epilepsy and {PNES}},
	url = {https://www.ncbi.nlm.nih.gov/pmc/articles/PMC7260313/},
	doi = {10.1186/s40708-020-00107-z},
	number = {1},
	urldate = {2023-09-03},
	journal = {Brain Informatics},
	author = {Ahmadi, Negar and Pei, Yulong and Carrette, Evelien and Aldenkamp, Albert P. and Pechenizkiy, Mykola},
	month = may,
	year = {2020},
	pmid = {32472244},
	pmcid = {PMC7260313},
	pages = {6},
}

@article{gedzelman_long-term_2014,
	title = {Long-term video {EEG} monitoring for diagnosis of psychogenic nonepileptic seizures},
	volume = {10},
	issn = {1176-6328},
	url = {https://www.ncbi.nlm.nih.gov/pmc/articles/PMC4206377/},
	doi = {10.2147/NDT.S49531},
	urldate = {2023-09-03},
	journal = {Neuropsychiatric Disease and Treatment},
	author = {Gedzelman, Evan R and LaRoche, Suzette M},
	month = oct,
	year = {2014},
	pmid = {25342907},
	pmcid = {PMC4206377},
	pages = {1979--1986},
}

@article{lafrance_jr_minimum_2013,
	title = {Minimum requirements for the diagnosis of psychogenic nonepileptic seizures: {A} staged approach},
	volume = {54},
	copyright = {Wiley Periodicals, Inc. © 2013 International League Against Epilepsy},
	issn = {1528-1167},
	shorttitle = {Minimum requirements for the diagnosis of psychogenic nonepileptic seizures},
	url = {https://onlinelibrary.wiley.com/doi/abs/10.1111/epi.12356},
	doi = {10.1111/epi.12356},
	language = {en},
	number = {11},
	urldate = {2023-09-03},
	journal = {Epilepsia},
	author = {LaFrance Jr., W. Curt and Baker, Gus A. and Duncan, Rod and Goldstein, Laura H. and Reuber, Markus},
	year = {2013},
	note = {\_eprint: https://onlinelibrary.wiley.com/doi/pdf/10.1111/epi.12356},
	pages = {2005--2018},
}

@article{brown_psychological_2016,
	title = {Psychological and psychiatric aspects of psychogenic non-epileptic seizures ({PNES}): {A} systematic review},
	volume = {45},
	issn = {0272-7358},
	shorttitle = {Psychological and psychiatric aspects of psychogenic non-epileptic seizures ({PNES})},
	url = {https://www.sciencedirect.com/science/article/pii/S0272735815301446},
	doi = {10.1016/j.cpr.2016.01.003},
	urldate = {2023-09-03},
	journal = {Clinical Psychology Review},
	author = {Brown, Richard J. and Reuber, Markus},
	month = apr,
	year = {2016},
	pages = {157--182},
}

@article{LancetDiagnosticChallengesEpilepsy,
	title = {Diagnostic challenges in epilepsy: seizure under-reporting and seizure detection},
	volume = {17},
	issn = {1474-4422},
	shorttitle = {Diagnostic challenges in epilepsy},
	url = {https://www.sciencedirect.com/science/article/pii/S1474442218300383},
	doi = {10.1016/S1474-4422(18)30038-3},
	number = {3},
	urldate = {2023-09-01},
	journal = {The Lancet Neurology},
	author = {Elger, Christian E and Hoppe, Christian},
	month = mar,
	year = {2018},
	pages = {279--288},
}

@article{seizureTypeClassification,
	title = {Current {Classification} of {Seizures} and {Epilepsies}: {Scope}, {Limitations} and {Recommendations} for {Future} {Action}},
	volume = {12},
	issn = {2168-8184},
	shorttitle = {Current {Classification} of {Seizures} and {Epilepsies}},
	url = {https://www.ncbi.nlm.nih.gov/pmc/articles/PMC7575300/},
	doi = {10.7759/cureus.10549},
	number = {9},
	urldate = {2023-09-01},
	journal = {Cureus},
	author = {Sarmast, Shah T and Abdullahi, Abba Musa and Jahan, Nusrat},
	pmid = {33101797},
	pmcid = {PMC7575300},
	pages = {e10549},
}

@article{keranen_distribution_1988,
	title = {Distribution of {Seizure} {Types} in an {Epileptic} {Population}},
	volume = {29},
	issn = {1528-1167},
	url = {https://onlinelibrary.wiley.com/doi/abs/10.1111/j.1528-1157.1988.tb05089.x},
	doi = {10.1111/j.1528-1157.1988.tb05089.x},
	language = {en},
	number = {1},
	urldate = {2023-09-02},
	journal = {Epilepsia},
	author = {Keränen, Tapani and Sillanpää, Matti and Riekkinen, Paavo J.},
	year = {1988},
	note = {\_eprint: https://onlinelibrary.wiley.com/doi/pdf/10.1111/j.1528-1157.1988.tb05089.x},
	pages = {1--7},
}

@article{mostCommonType,
	series = {Common {Neurologic} {Disorders}},
	title = {Seizure {Disorders}},
	volume = {93},
	issn = {0025-7125},
	url = {https://www.sciencedirect.com/science/article/pii/S0025712508001557},
	doi = {10.1016/j.mcna.2008.10.001},
	number = {2},
	urldate = {2023-09-01},
	journal = {Medical Clinics of North America},
	author = {Schachter, Steven C.},
	month = mar,
	year = {2009},
	pages = {343--351},
}

\end{document}